\documentclass[11pt,a4paper]{article}
\usepackage[hyperref]{acl2021}
\usepackage{times}
\usepackage{latexsym}

\usepackage{xspace}
\usepackage{graphicx}
\usepackage{caption}
\usepackage{subcaption}
\usepackage{times}
\usepackage{url}
\usepackage{latexsym}
\usepackage{amsmath}
\usepackage{lipsum}
\usepackage{multirow}
\usepackage{booktabs}
\usepackage{enumerate}
\usepackage{dblfloatfix}
\usepackage{cleveref}
\usepackage{microtype}

\usepackage{microtype}
\usepackage{enumitem}

\setlist[itemize]{noitemsep, topsep=0pt} 
\aclfinalcopy % Uncomment this line for the final submission

\newcommand{\comment}[1]{}

\crefname{section}{\S}{\S\S}
\Crefname{section}{\S}{\S\S}

\newcommand{\corpusname}{ILDC\xspace}

\newcommand{\corpuslarge}{\ensuremath{\text{\corpusname}_{\text{multi}}}\xspace}
\newcommand{\corpussmall}{\ensuremath{\text{\corpusname}_{\text{single}}}\xspace}
\newcommand{\corpusexpert}{\ensuremath{\text{\corpusname}_{\text{expert}}}\xspace}

\title{ILDC for CJPE: Indian Legal Documents Corpus for Court Judgment Prediction and Explanation}
\author{Vijit Malik$^{1}$ \qquad Rishabh Sanjay$^{1}$ \qquad Shubham Kumar Nigam$^{1}$\\
\textbf{Kripa Ghosh}$^{2}$ \qquad \textbf{Shouvik Kumar  Guha}$^{3}$ \qquad \textbf{Arnab Bhattacharya}$^{1}$\\
\textbf{Ashutosh Modi}$^{1}$ \\
 $^{1}$Indian Institute of Technology Kanpur (IIT-K) \\
 $^{2}$Indian Institute of Science Education and Research  Kolkata (IISER-K)\\
 $^{3}$West Bengal National University of Juridical Sciences (WBNUJS)\\
 \texttt{\{vijitvm,rsan,sknigam\}@iitk.ac.in} \\ 
 \texttt{kripaghosh@iiserkol.ac.in} \qquad
 \texttt{shouvikkumarguha@nujs.edu} \\
 \texttt{\{arnabb,ashutoshm\}@cse.iitk.ac.in}
}%author

\date{}

\begin{document}
\maketitle

%\vspace{10mm}

% \vspace{-20cm}
\begin{abstract}
\vspace{-3mm}

An automated system that could assist a judge in predicting the outcome of a case would help expedite the judicial process. For such a system to be practically useful, predictions by the system should be explainable. To promote research in developing such a system, we introduce \textit{ILDC (Indian Legal Documents Corpus)}. ILDC is a large corpus of 35k Indian Supreme Court cases annotated with original court decisions. A portion of the corpus (a separate test set) is annotated with gold standard explanations by legal experts. Based on ILDC, we propose the task of Court Judgment Prediction and Explanation (CJPE). The task requires an automated system to predict an explainable outcome of a case.  We experiment with a battery of baseline models for case predictions and propose a hierarchical occlusion based model for explainability. Our best prediction model has an accuracy of 78\% versus 94\% for human legal experts, pointing towards the complexity of the prediction task. The analysis of explanations by the proposed algorithm reveals a significant difference in the point of view of the algorithm and legal experts for explaining the judgments, pointing towards scope for future research.

\end{abstract}

%\vspace{-3mm}

\section{Introduction}
\label{sec:intro}
%\vspace{-2mm}
%\AM{I think we should change the title of the paper a bit as it has two judgment word and looks a bit weird, one possibility: ``Court Judgment Prediction and Explanation (CJPE) on Unstructured Supreme Court Case Documents"} \KG{Supreme Court Decisions? The Supreme Court can have many types of documents e.g. evidence, minutes (roznama) etc. We are looking only at decisions.} \AB{Simply Supreme Court Cases?} \AM{I agree with AB}

\noindent In many of the highly populated countries like India, there is a vast number of pending backlog of legal cases that impede the judicial process \cite{backlog-cases2019}. %\footnote{\url{https://tinyurl.com/y7w9hrfl}}. 
The backlog is due to multiple factors, including the unavailability of competent judges.  %suited for the case at hand. 
Therefore, a system capable of assisting a judge by \emph{suggesting} the outcome of an ongoing court case is likely to be useful for expediting the judicial process.
However, an automated decision system is not tenable in law unless it is well explained in terms of how humans understand the legal process. 
Hence, it is necessary to \emph{explain} the suggestion.
In other words, we would like such a system to predict not only \emph{what} should be the final decision of a court case but also \emph{how} one arrives at that decision. In this paper, we introduce \textsc{Indian Legal Documents Corpus (ILDC)} intending to promote research in developing a system that could \textit{assist} in legal case judgment prediction in an explainable way. ILDC is a corpus of case proceedings from the Supreme Court of India (SCI) that are annotated with original court decisions. A portion of ILDC (i.e., a separate test set) is additionally annotated with gold standard judgment decision explanations by legal experts to evaluate how well the judgment prediction algorithms explain themselves.  
%We would like the prediction algorithms to be explainable without requiring any  that are able to explain themselves, the test set of the corpus is additionally annotated with gold standard decision explanations by legal experts. %To the best of our knowledge, ILDC is the first legal corpus with explanation annotations for the decisions. For the corpus creation, we involved legal experts who annotated the documents. They also help understand the thought process when examining a case; this is important from the modeling perspective.

Based on ILDC, we propose a new task: \textsc{Court Judgment Prediction and Explanation (CJPE)}. This task aims to predict the \emph{final decision} given all the \emph{facts and arguments} of the case and provide an explanation for the predicted decision. The decision can be either \emph{allowed}, which indicates ruling in favor of the appellant/petitioner, or \emph{dismissed}, which indicates a ruling in favor of the respondent. The explanations in the CJPE task refer to sentences/phrases in the case description that best justify the final decision. Since, we are addressing mainly the SCI cases, one might argue that the usefulness of the task may be limited since, the legislative provisions can always change with time. However, the legal principles of how to apply a given law to a given set of facts remain constant for prolonged periods.

Judgment prediction and explanation in the CJPE task are far more challenging than a standard text-classification task for multiple reasons. Firstly, the legal court case documents (especially in Indian context) are unstructured and are usually quite long, verbose, and noisy. There is no easy way of extracting and directly using the facts and arguments. Secondly, the domain-specific lexicon used in court cases makes models pre-trained on generally available texts ineffective on such documents. Consequently, the standard models need to be adapted to the legal domain for the proposed judgment prediction on court cases. Thirdly, explaining prediction in legal documents is considerably more challenging as it requires understanding the facts, following the arguments and applying legal rules, and principles to arrive at the final decision. \\
\noindent Our main contributions can be summarized as:\\
1. We create a new corpus, \textsc{Indian Legal Documents Corpus (ILDC)}, annotated with court decisions. A portion of the corpus (i.e. a separate test set) is additionally annotated with explanations corresponding to the court decisions. We perform detailed case studies on the corpus to understand differences in prediction and explanation annotations by legal experts, indicative of the computational challenges of modeling the data. \\
2. We introduce a new task, \textsc{Court Judgment Prediction and Explanation (CJPE)}, with the two sub-tasks: (a)~Court Judgment Prediction (CJP) and (b)~Explanation of the Prediction. While CJP is not a novel task per se; however, in combination with the explanation part, the CJPE task is new. Moreover, the requirement for explanations also puts restrictions on the type of techniques that could be tried for CJP. In the CJPE task, gold explanations are not provided in the train set; the task expects that the trained algorithms should explain the predictions without requiring additional information in the form of annotations during training.  \\
3. We develop a battery of baseline models for the CJPE task. We perform extensive experimentation with state-of-the-art machine learning algorithms for the judgment prediction task. We develop a new method for explaining machine predictions since none of the existing methods could be readily applied in our setting. We compare model explainability results with annotations by legal experts, showing significant differences between the point of view of algorithms and experts.

ILDC is introduced to promote the development of a system/models that will \textit{augment humans and not replace} them. We have covered the ethical considerations in the paper. Nevertheless, the community needs to pursue more research in this regard to fully understand the unforeseen social implications of such models. This paper takes initial steps by introducing the corpus and baseline models to the community. Moreover, we plan to continue to grow, revise and upgrade ILDC. We release the ILDC and code for the prediction and explanation models via GitHub\footnote{\url{https://github.com/Exploration-Lab/CJPE}}. %Please check out the GitHub repository for a detailed analysis of our methods and case studies upon the expert annotations.
 
%\vspace{-1mm}

\section{Related Work} \label{sec:related}
\vspace{-1.5mm}

%\AM{Related work doesn't look polished, it needs work.}

%\subsection{Legal Judgment Prediction (LJP) and Charge Prediction} \label{sec:related-ljp}
%\vspace{-0.3cm}
%Machine learning has been used in multiple tasks in the legal domain like Prior Case Retrieval, Summarization, Catchphrase Extraction, Crime Classification, Judgment Prediction etc.
\noindent There has been extensive research on legal domain text, and various corpora and tasks have been proposed e.g., prior case retrieval \cite{jackson2003information}, summarization \cite{Tran2019,bhattacharya2019comparative}, catchphrase extraction \cite{galgani2012towards}, crime classification \cite{Wang2019}, and judgment prediction \cite{Zhong_Wang_Tu_Zhang_Liu_Sun_2020}. 

\noindent\textbf{Why ILDC?} The task of  Legal Judgment Prediction (LJP) and its corresponding corpora \citep{chalkidis-etal-2019-neural, Zhong_Wang_Tu_Zhang_Liu_Sun_2020, Yang2019, xiao2018cail2018} are related to our setting. In the LJP task, given the \emph{facts} of a case, \emph{violations}, \emph{charges} (e.g., theft) and \emph{terms of penalty} are predicted. However, the ILDC and the CJPE task introduced in this paper differ from the existing LJP corpora and task in multiple ways. Firstly, we require prediction algorithms to explain the decisions in the CJPE task, to evaluate the explanations we provide a separate test set annotated with gold explanations. Secondly, in the LJP task, typically, the facts of a case are explicitly provided. However, in our case, only unannotated unstructured documents are provided. %, and facts, charges are not annotated. 
ILDC addresses a more realistic/practical setting, and consequently, CJPE is a much more challenging task. Moreover, the bare facts do not form the judgment premise of a case since facts are subject to interpretations. A court case description, in practice, has other vital aspects like {\it Ruling by Lower Court}, {\it Arguments}, {\it Statutes}, {\it Precedents}, and {\it Ratio of the decision} \cite{bhattacharya2019identification} that are instrumental in decision making by the judge(s). Unlike LJP, we consider (along with the facts) the entire case (except the judgment), and we predict the judgment only. Work by \citet{Strickson2020} comes close to our setting, where the authors prepared the test set on UK court cases by removing the final decision from rulings and employed classical machine learning models. Thirdly, to the best of our knowledge, we are the first to create the largest legal corpus ($34,816$ documents) for the Indian setting. It is important because India has roots in the common law system and case decisions are not strictly as per the statute law, with the judiciary having the discretion to interpret their version of the legal provisions as applicable to the case at hand; % or even exercise residuary power entrusted with the apex judiciary to ensure justice; 
this can sometimes make the decision process subjective. Fourth, we do not focus on any particular class of cases (e.g., criminal, civil) but address publicly available generic SCI case documents. %\SN{Thirdly, instead of filtered case domain, we included all publicly available cases of the Supreme Court of India.}

\citet{xiao2018cail2018} released the Chinese AI and Law challenge dataset (CAIL2018) in Chinese for judgment prediction, that contains more than $2.68$ million \textit{criminal cases} published by the Supreme People’s Court of China.  \citet{chalkidis-etal-2019-neural} released an English legal judgment prediction dataset, containing $11,478$ cases from the European Court of Human Rights (ECHR). It contains facts, articles violated (if any), and an importance score for each case. ILDC contrasts with the existing LJP corpora, where mainly the civil law system and cases are considered. Though the proposed corpus focuses on Indian cases, our analysis reveals (\cref{sec:analysis-anno}) that the language used in the cases is quite challenging to process computationally and provides a good playground for developing realistic legal text understanding systems.     

Several different approaches and corpora have been proposed for the LJP task. \newcite{chalkidis-etal-2019-neural} proposed a hierarchical version of BERT \cite{devlin-etal-2019-bert} to alleviate BERT's input token count limitation for the LJP task. %They released an English language dataset of $11$k cases from the European Court of Human Rights. It contains facts, articles violated (if any), and an importance score for each case. 
\newcite{Yang2019} applied Multi-Perspective Bi-Feedback Network for predicting the relevant law articles, charges, and terms of penalty on Chinese AI and Law challenge (CAIL2018) datasets.  %that contain fact descriptions, applicable law articles, charges, and the terms of penalty.  
\newcite{xu-etal-2020-distinguish} proposed a system for distinguishing confusing law articles in the LJP task. \newcite{LJPTopologicalLearning} applied topological multi-task learning on a directed acyclic graph to predict charges like theft, traffic violation, intentional homicide on three Chinese datasets (CJO, PKU, and
CAIL). \newcite{ChargePredictionLegalBasis} proposed an attention-based model to predict the charges given the facts of the case along with the relevant articles on a dataset of Criminal Law of the People's Republic of China. \newcite{ChargePredictionFewShot} used an attribute-attentive model in a few-shot setup for charge prediction from facts of the case. \newcite{long2019automatic} predicts the decision of the case using a Legal Reading Comprehension technique on a Chinese dataset. \newcite{chen-etal-2019-charge} used a deep gating network for prison term prediction, given the facts and charges on a dataset constructed from documents of the Supreme People's Court
of China. \newcite{Aletras2016} used linear SVM to predict violations from facts on European Court of Human Rights cases. \newcite{sulea2017} used SVM in the LJP task on French Supreme Court cases. \newcite{Katz2017} presented a random forest model to predict the ``Reverse'', ``Affirm'', and ``Other'' decisions of US Supreme Court judges. We also experiment with some of these models as baselines for the CJPE task (\cref{sec:cjpe}). 

% \vspace{-3mm}
\begin{table}
\small
%\vspace{-6mm}
% \usepackage{multirow}
\begin{tabular}{|c|c|c|c|}
\hline
	\multirow{2}{*}{\textbf{\begin{tabular}[c]{@{}c@{}}Corpus\\ (Avg. tokens)\end{tabular}}} &
	\multicolumn{3}{c|}{\textbf{\begin{tabular}[c]{@{}c@{}}Number of docs \\ (Accepted Class \%)\end{tabular}}} \\
\cline{2-4} 
& \textbf{Train}    & \textbf{Validation} & \textbf{Test}                                                              \\ \hline
	\textbf{\begin{tabular}[c]{@{}c@{}}\corpuslarge\\ (3231)\end{tabular}} &
	\begin{tabular}[c]{@{}c@{}}32305\\ (41.43\%)\end{tabular} &
		\multirow{2}{*}{\begin{tabular}[c]{@{}c@{}}994\\ (50\%)\end{tabular}} &
		\multirow{2}{*}{\begin{tabular}[c]{@{}c@{}}1517\\ (50.23\%)\end{tabular}} \\
\cline{1-2}
\textbf{\begin{tabular}[c]{@{}c@{}}\corpussmall\\ (3884)\end{tabular}}                     & \begin{tabular}[c]{@{}c@{}}5082\\ (38.08\%)\end{tabular}  &                                                                        &                                                                            \\ \hline
\textbf{\begin{tabular}[c]{@{}c@{}}\corpusexpert\\ (2894)\end{tabular}}                   & \multicolumn{3}{c|}{56 (51.78\%)}                                                                                                                                                                               \\ \hline
\end{tabular}
%\vspace{-3mm}
\caption{ILDC Statistics}% \AB{extremely strange \LaTeX{} coding -- will clean up later}}
\label{tab:ildc}
%\vspace{-7.2mm}
\end{table}

Explainability in a system is of paramount importance in the legal domain. \newcite{Zhong_Wang_Tu_Zhang_Liu_Sun_2020} presented a QA based model using reinforcement learning for {\it explainable} LJP task on three Chinese datasets (CJO, PKU, and
CAIL). The model aims to predict the appropriate crime by asking relevant questions related to the facts of the case. \newcite{jiang-interpretableChargePrediction} used a rationale augmented classification model for the charge prediction task. The model selects as rationale the relevant textual portions in the fact description. \newcite{ye-interpretableChargePrediction} used label-conditioned Seq2Seq model for charge prediction on Chinese legal documents, and the interpretation comprise the selection of the relevant rationales in the text for the charge. We develop an explainability model based on the occlusion method  (\cref{sec:explanation-model}). 
\vspace{-2mm}

\section{Indian Legal Document Corpus} \label{sec:ildc} 

\noindent In this paper, we introduce the \textsc{Indian Legal Documents Corpus (ILDC)}, a collection of case proceedings (in the English language) from the Supreme Court of India (SCI). For a case filed at the SCI, a decision (``accepted'' v/s ``rejected'') is taken between the appellant/petitioner versus the respondent by a judge while taking into account the {\it facts of the case}, {\it ruling by lower Court(s)}, if any, {\it arguments}, {\it statutes}, and {\it precedents}. For every case filed in the Supreme Court of India (SCI), the judge (or a bench) decides on whether the claim(s) filed by the appellant/petitioner against the respondent should be ``accepted'' or ``rejected''. The decision is relative to the appellant. In ILDC, each of the case proceeding document is labeled with the original decision made by the judge(s) of the SCI, which serve as the gold labels. In addition to the ground truth decision, a separate test set documents are annotated (by legal experts) with explanations that led to the decision. The explanations annotations are ranked in the order of importance. %An example of a case document from ILDC is given in Table \ref{tab:case_description} in Appendix \ref{supp:case-study}. \SN{We have removed the example case from the appendix, so need to remove this line from here.} 

\noindent\textbf{ILDC Creation.} We extracted all the publicly available SCI\footnote{Although IndianKanoon includes lower court cases as well, they do not have a common structural format and  many of the case documents in lower courts may be in a regional Indian language. Hence, for now we only use SCI documents.} case proceedings from the year 1947 to April 2020 from the website: \url{https://indiankanoon.org}. Case proceedings are unstructured documents and have different formats and sizes, have spelling mistakes (since these are typed during the court hearing), making it challenging to (pre-)process. We used regular expressions to remove the noisy text and meta-information (e.g., initial portions of the document containing case number, judge name, dates, and other meta information) from the proceedings. In practice, as pointed by the legal experts, the judge deciding the case and other meta information influence the final decision. In SCI case proceedings, the decisions are written towards the end of the document. These end section(s) directly stating the decision have been deleted from the documents in ILDC since that is what we aim to predict. Each case's actual decision label has been extracted from the deleted end sections of the proceeding using regular expressions. Another challenge with SCI case proceedings is the presence of cases with multiple petitions where, in a single case, multiple petitions have been filed by the appellant leading to multiple decisions. Consequently, we divided ILDC documents into two sets. The first set, called \corpussmall, either have documents where there is a single petition (and, thus, a single decision) or multiple petitions, but the decisions are the same across all those petitions. The second set, called \corpuslarge, is a superset of \corpussmall and has multiple appeals leading to different decisions. Predicting multiple different decisions for cases with multiple appeals is significantly challenging. In this paper, we do not develop any baseline computational models for this setting; we plan to address this in future work. For the computational models for the CJPE task, in the case of \corpuslarge, even if a single appeal was accepted in the case having multiple appeals/petitions, we assigned it the label as accepted.  Table \ref{tab:ildc} shows the corpus statistics for ILDC. Note that the validation and test sets are the same for both \corpuslarge and \corpussmall.

\noindent\textbf{Temporal Aspect.} The corpus is randomly divided into train, validation, and test sets, with the restriction that validation and test sets should be balanced w.r.t. the decisions. The division into train, development, and test set was not based on any temporal consideration or stratification because the system's objective that may eventually emerge from the project is not meant to be limited to any particular law(s), nor focused on any particular period of time. On the contrary, the aim is to identify standard features of judgments pronounced in relation to various legislation by different judges and across different temporal phases, to be able to use the said features to decipher the judicial decision-making process and successfully predict the nature of the order finally pronounced by the court given a set of facts and legal arguments. While there would be a degree of subjectivity involved, given the difference in the thoughts and interpretations adopted by different judges, such differences are also found between two judges who are contemporaries of each other, as much as between two judges who have pronounced judgments on similar matters across a gap of decades. The focus is, therefore, to develop a system that would be equally successful in predicting the outcome of a judgment given the law that had been in vogue twenty years back, as it would in relation to the law that is currently in practice. The validity and efficacy of the system can therefore be equally tested by applying it to cases from years back, as to cases from a more recent period. In fact, if the system cannot be temporally independent, and remains limited to only successful prediction of contemporary judgments, then it is likely to fail any test of application because by the time the final version of the system can be ready for practical applications on a large scale, the laws might get amended or replaced, and therefore, the judgments that would subsequently be rendered by the court might be as different from one pronounced today, as the latter might differ from one pronounced in the twentieth century. Not acknowledging time as a factor during data sample choice, therefore, appears to be the prudent step in this case, especially given the exponential rate at which legislation is getting amended today, as well as the fast-paced growth of technological development. 
\begin{table*}[t]
\parbox{.32\linewidth}{
\vspace{-5mm}
\small
\begin{center}
\begin{tabular}{|c|c|}
\hline
%\caption{Annotators' prediction accuracy.}
\textbf{Expert} & \textbf{Accuracy (\%)} \\ \hline
\textbf{Expert 1} & 94.64                  \\ \hline
\textbf{Expert 2} & 91.07                  \\ \hline
\textbf{Expert 3} & 98.21                  \\ \hline
\textbf{Expert 4} & 89.28                  \\ \hline
\textbf{Expert 5} & 96.43                  \\ \hline
\end{tabular}
%\vspace{-2.5mm}
\caption{Annotators' accuracy.}
\vspace{-5mm}
\label{tab:user_accuracy}
%\vspace{-10mm}
\end{center}
}
% \parbox{.03\linewidth}{
% \begin{tabular}{|c|c|}
% \end{tabular}
% }
\parbox{.65\linewidth}{
\small
\centering
%\vspace{1mm}
\begin{tabular}{|c|c|c|c|c|c|}
\hline
%\multirow{2}{*}{\textbf{Judgment Accuracy}} & \multicolumn{5}{c|}{\textbf{Reference text}}                                            \\ \cline{2-6} 
{\bf Agreement (\%)}                                             & \textbf{Expert 1} & \textbf{Expert 2} & \textbf{Expert 3} & \textbf{Expert 4} & \textbf{Expert 5} \\ \hline
\textbf{Expert 1}                              & 100.0	               & 87.5	          & 94.6	       &    85.7	     &    89.3          \\ \hline
\textbf{Expert 2}                              & 87.5	       & 100.0	              & 92.9	       &    87.5	     & 91.1          \\ \hline
\textbf{Expert 3}                              & 94.6	       & 92.9	       & 100.0               &    91.1	     & 94.6          \\ \hline
\textbf{Expert 4}                              & 85.7	       & 87.5	       & 91.1	           & 100.0	             & 89.3          \\ \hline
\textbf{Expert 5}                              & 89.3	       & 91.1	       & 94.6	           & 89.3           & 100.0              \\ \hline
\end{tabular}
%\vspace{-2.5mm}
\caption{Pairwise inter-annotator agreement for judgment prediction.}
%\vspace{-5mm}
\label{tab:agreement-judgment}
}
\end{table*}

\noindent \textbf{Legal Expert Annotations.} In our case, the legal expert team consisted of a law professor and his students at a reputed national law school. We took a set of $56$ documents (\corpusexpert) from the test set, and these were given to $5$ legal experts.  Experts were requested to (i)~predict the judgment, and (ii)~mark the sentences that they think are explanations for their judgment. Each document was annotated by all the 5 experts (in isolation) using the WebAnno framework \cite{de2016web}. The annotators could assign ranks to the sentences selected as explanations; a higher rank indicates more importance for the final judgment. The rationale for rank assignment to the sentences is as follows. {\it Rank 1} was given to sentences immediately leading to the decision. {\it Rank 2} was assigned to sentences that contributed to the decision. {\it Rank 3} was given to sentences indicative of the disagreement of the current court with a lower court/tribunal decision. Sentences containing the facts of the case, not immediately, leading to decision making, but are essential for the case were assigned {\it Rank 4 (or lower)}. Note in practice, only a small set of sentences of a document were assigned a rank. Although documents were annotated with explanations in order of ranks, we did not have a similar mechanism in our automated explainability models. From the machine learning perspective, this is a very challenging task, and to the best of our knowledge, none of the state-of-the-art explainability models are capable of doing this. Annotation of explanations is a very specialized, time-consuming, and laborious effort. In the current version of ILDC we provide explanation annotations to only a small portion of the test set, this is for evaluating prediction algorithms for the explainability aspect. Even this small set of documents is enough to highlight the difference between the ML-based explainability methods and how a legal expert would explain a decision (\cref{sec:explainabilitymodelVsAnnotators}). Nevertheless, we plan to continue to grow the corpus by adding more explainability annotations and other types of annotations. Moreover, we plan to include lower courts like Indian High Court cases and tribunal cases. The corpus provides new research avenues to be explored by the community. 

\noindent\textbf{Fairness and Bias.} While creating the corpus, we took all possible steps to mitigate any biases that might creep in. We have not made any specific choice with regard to any specific law or any category of cases, i.e., the sampling of cases was completely random. As explained earlier, we took care of the temporal aspect. Importantly, the names of the judge(s), appellants, petitioners, etc., were anonymized in the documents so that no inherent bias regarding these creeps in. The anonymization with respect to judge names is necessary as legal experts pointed out that a judge's identity can sometimes be a strong indicator of the case outcome. It is noteworthy that according to the legal experts if we had not done the same, we could have had higher prediction accuracy. The subjectivity associated with judicial decision-making may also be controlled in this way since the system focuses on how consideration of the facts and applicable law are supposed to determine the outcome of the cases, instead of any individual bias on the judge's part. We also address the ethical concerns in the end.

\vspace{-2mm}

\section{Annotation Analysis} \label{sec:annoanalysis}
\vspace{-1.5mm}
% \begin{figure}[!h]
% %\vspace{-4mm}
%         \centering
%          \vspace*{-4mm}
%          \includegraphics[width=\linewidth]{images/ROUGE-L.pdf}
%          \vspace*{-13mm}
%         \caption{Explanation agreement among the annotators}% (users) \AB{why are we mentioning ``users''?}: ROUGE-L.}
%         \label{fig:users-rougeLOverlapMin}
%         \vspace{-3.5mm}
% \end{figure}

\noindent We performed a detailed analysis of case predictions and the explanations annotations. With assistance from a legal expert, we also performed detailed studies for some court cases to understand the task's complexity and possible reasons for deviations between the annotators.  %(\cref{supp:case-study})

\vspace{-2.5mm}
\subsection{Case Judgment Accuracy} \label{sec:analysis-pred}
%\noindent {\bf Annotators:} 
We computed the case judgment accuracy of the annotators with respect to original decisions by judges of SCI. The results are shown in Table \ref{tab:user_accuracy}. Though the values are high, none of these are 100\%. The accuracy indicates that no annotator agrees with the original judgment in all the cases. This possibly depicts the subjectivity in the legal domain with regard to decision making. The subjectivity aspect has also been observed in other tasks that involve human decision-making, e.g., sentiment and emotion analysis. We performed detailed case studies with the help of experts to further probe into this difference in judgment. Due to space limitations, we are not able to present the studies here; please refer to \cref{supp:case-prediction} and GitHub repository %\SN{Is this appendix reference is still relevant?} 
for details. To summarize, the study indicated that the sources of confusion are mainly due to differences in linguistic interpretation (by the annotators) of the legal language given in the case document.  

\subsection{Inter-Annotator Agreements} \label{sec:analysis-anno}

\noindent {\bf Agreement in the judgment prediction:} For the {\it quantitative evaluation}, we calculate pair-wise agreement between the annotators as shown in Table \ref{tab:agreement-judgment}. The highest agreement $(94.6\%)$ is between Experts 1-3 and 3-5. We also calculate Fleiss' kappa \cite{FleissKappa} as $0.820$, among all the five annotators, which indicates high agreement. 

\begin{figure}[t]%[!h]
\vspace{4.5mm}
        \centering
         %\vspace*{-4.5mm}
         \includegraphics[width=\linewidth]{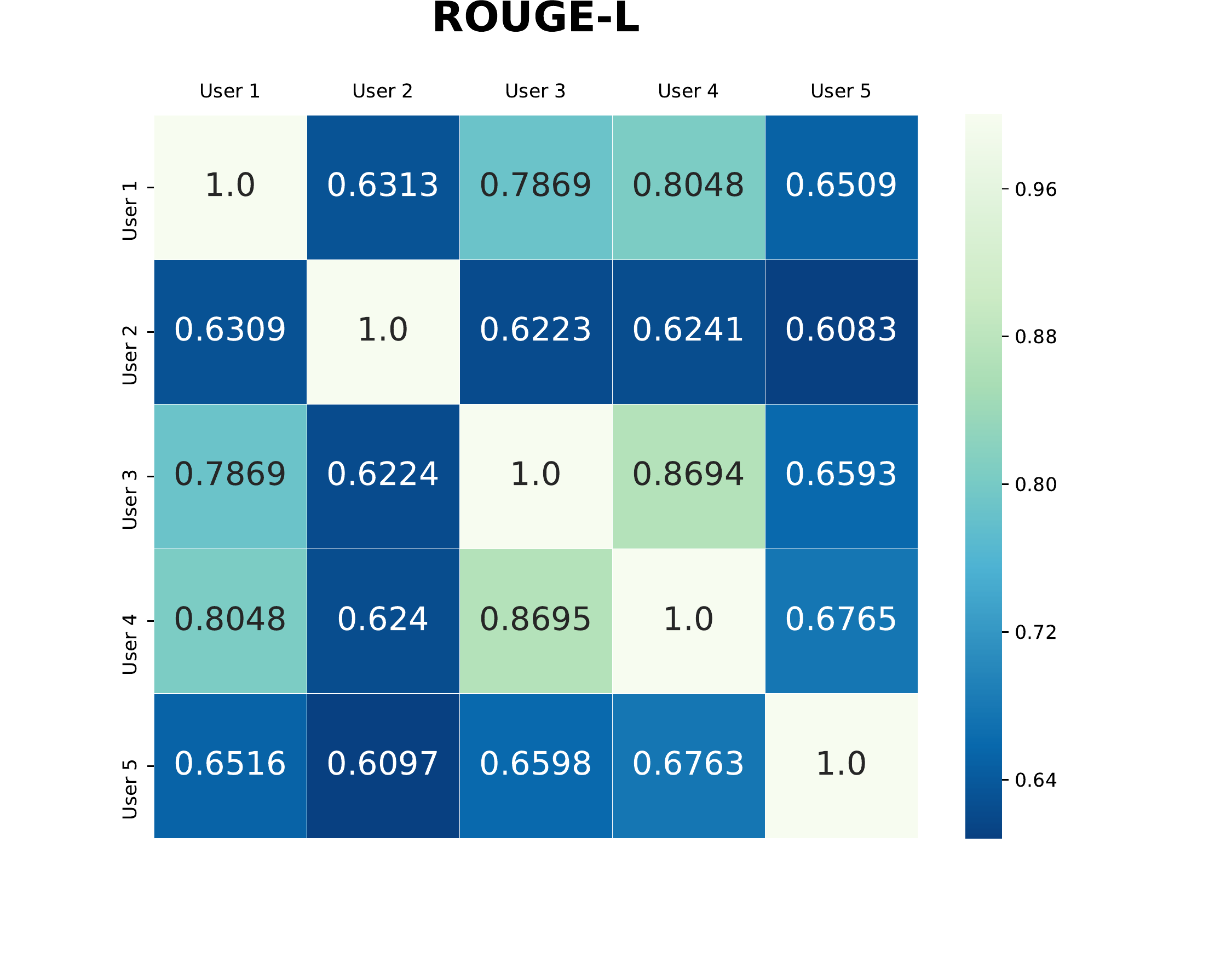}
         \vspace*{-13mm}
        \caption{Explanation agreement among the annotators}% (users) \AB{why are we mentioning ``users''?}: ROUGE-L.}
        \label{fig:users-rougeLOverlapMin}
        \vspace{-4.5mm}
\end{figure}

\noindent {\bf Agreement in the explanation:}
There are no standard metrics for evaluating annotator agreements for textual annotations. For {\it quantitative evaluation} of agreements among the annotators for explanations, we took inspiration from machine translation community and used metrics like ROUGE-L, ROUGE-1, ROUGE-2 \cite{lin2004rouge}, BLEU \cite{papineni2002bleu} (unigram and bigram averaging), METEOR \cite{lavie2007meteor}, Jaccard Similarity, %\footnote{\url{https://en.wikipedia.org/wiki/Jaccard_index}}, 
 Overlap Maximum and Overlap Minimum\footnote{Overlap Max: Size of the intersection divided by the maximum size out of the two sample sets that are being compared. Overlap Min: Size of the intersection divided by the minimum size out of the two sample sets that are being compared}. The result for ROUGE-L (averaged out over all documents)\footnote{Due to space constraints we are not able to show heatmaps corresponding to other metrics but they showed similar trends. For the heatmaps for other metrics please refer to our GitHub repository.} is shown in Figure \ref{fig:users-rougeLOverlapMin}.  %Please refer to Appendix \ref{supp:interannotatoragreement} to see heat-maps corresponding to other metrics\RS{Sir as we have removed the heatmaps from the appendix, so do we need to remove the metrics from here?}. 
 The highest overlap across all the metrics is observed between Expert~3 and Expert~4. The highest value $(0.9129)$ is between Expert~2 and Expert~4 for Overlap-Min. We also performed a {\it qualitative evaluation} of the agreements in the explanations. %(Appendix \ref{supp:case-explanations-annotations}). 
We observed that Expert~1, Expert~3, and Expert~4 consider holistic reasoning for the decision. They look at both Substantive (sections applicable) and Procedural (about the jurisdiction of a lower court) aspects of the case. The differences among them are largely due to consideration/non-consideration of the factual sentences. On the other hand, Expert~2 and Expert~5 often use bare-minimum reasoning leading to the final judgment instead of looking at the exhaustive set of reasons and did not always cover both Substantive and Procedural aspects of the case. 

% \begin{figure}[t]%[h]
% %\vspace{-4mm}
%         \centering
%          %\vspace*{-5mm}
%          \includegraphics[width=\linewidth]{images/ROUGE-L.pdf}
%          \vspace*{-13mm}
%         \caption{Explanation agreement among the annotators}% (users) \AB{why are we mentioning ``users''?}: ROUGE-L.}
%         \label{fig:users-rougeLOverlapMin}
%         \vspace{-11mm}
% \end{figure}

Analysis of annotations gives insights into the inherent complexity and subjectivity of the task. Legal proceedings are long, verbose, often challenging to comprehend, and exhibit interesting (and computationally challenging) linguistic phenomena. For example, in a case numbered ``1962\_47'' (\cref{supp:case-prediction}), sentence 17 of the case appears to refer to the Supreme Court having accepted a previous appeal for which a review has been requested (i.e., the current appeal). This amounted to the fact that the court actually rejected the present appeal while accepting the previous one. Such intricacies can confuse even legal experts. %(\cref{supp:case-prediction}).
\vspace{-2mm}

\section{CJPE Task}
\label{sec:cjpe}

\noindent Given a case proceeding from the SCI, the task of \textsc{Court Judgment Prediction and Explanation (CJPE)} is to automatically predict the decision for the case (with respect to the appellant) and provide the explanation for the decision. We address the CJPE task via two sub-tasks in the following sequence: Prediction and Explanation. 

\noindent \textbf{Prediction}: Given a case proceeding $D$, the task is to predict the decision $y \in \{0,1\}$, where the label $1$ corresponds to the acceptance of the appeal/petition of the appellant/petitioner. 

\noindent \textbf{Explanation}: Given the case proceeding and the predicted decision for the case, the task is to explain the decision by predicting important sentences that lead to the decision. Annotated explanations are not provided during training; the rationale is that a model learned for prediction should explain the decision without explicit training on explanations, since explanation annotations are difficult to obtain. %

\subsection{Case Decision Prediction} \label{sec:models-prediction}
%\vspace{-1mm}
\noindent ILDC documents are long and have specialized vocabulary compared to typical corpora used for training text classification models and language models. We initially experimented with non-neural models based on text features (e.g., n-grams, tf-idf, word based features, and syntactic features) and existing pre-trained models (e.g., pre-trained word embeddings based models, transformers), but none of them were better than a random classifier. Consequently, we retrained/fine-tuned/developed neural models for our setting. In particular, we ran a battery of experiments and came up with four different types of models: classical models, sequential models, transformer models, and hierarchical transformer models. Table \ref{tab:results} summarizes the performance of different models. Due to space constraints, we are not able to describe each of the models here. We give a very detailed description of model implementations in \cref{supp:model-details}.

\noindent\textbf{Classical Models:} We considered classical ML models like word/sentence embedding based  Logistic Regression, SVM, and Random Forest. We also tried prediction with summarized legal \citep{bhattacharya2019comparative} documents; however, these resulted in a classifier no better than random classifier. As shown in Table \ref{tab:results}, classical models did not perform so well. However, model based on Doc2vec embeddings had similar performance as sequential models. 

We extensively experimented with dividing documents into chunks and training the model using each of the chunks separately. We empirically determined that sequential and transformer-based models performed the best on the validation set using the last 512 tokens\footnote{length of 512 was partly influenced by the maximum input token limit of BERT} of the document. Intuitively, this makes sense since the last parts of case proceedings usually contain the main information about the case and the rationale behind the judgment. We also experimented with different sections of a document, and we observed last 512 tokens gave the best performance.

\begin{table}[!h]
% \vspace{-2mm}
%\begin{center}
%\small
% \tiny
%\hski-2.1cm
\resizebox{\columnwidth}{!}{
\begin{tabular}{|l|l|l|l|l|}
\hline
%%%%%%%%%%%%%%%%%%%%%%%%%%%%%%%%%%%%%%%%%%%%%%%%%%%%
\textbf{Model} & \textbf{\begin{tabular}[c]{@{}l@{}}Macro \\ Precision\\ (\%)\end{tabular}} & \textbf{\begin{tabular}[c]{@{}l@{}}Macro \\ Recall \\(\%)\end{tabular}} & \textbf{\begin{tabular}[c]{@{}l@{}}Macro\\ F1\\(\%)\end{tabular}} & \textbf{\begin{tabular}[c]{@{}l@{}}Accuracy \\ (\%)\end{tabular}} \\ \hline
%%%%%%%%%%%%%%%%%%%%%%%%%%%%%%%%%%%%%%%%%%%%%%%%%%%%
\multicolumn{5}{|l|}{\textbf{Classical Models on \corpuslarge train set}} \\ \hline
\textbf{Doc2Vec + LR} & 63.03 & 61.00 & \textbf{62.00} & 60.91\\\hline
Sent2vec + LR & 57.19 & 55.55 & 56.36 & 55.44\\ \hline
%%%%%%%%%%%%%%%%%%%%%%%%%%%%%%%%%%%%%%%%%%%%%%%%%%%%
\multicolumn{5}{|l|}{\textbf{Sequential Models on \corpuslarge train set}} \\ \hline
Sent2vec + BiGRU + att. & 60.98 & 58.40 & 59.66 & 58.31 \\ \hline
Doc2vec + BiGRU + att. & 57.18 & 56.03 & 56.60 & 57.44 \\ \hline 
\textbf{GloVe + BiGRU + att.} & 68.26 & 60.87 & \textbf{64.35} & 60.75 \\ \hline
HAN & 59.96 & 59.57 & 59.77 & 59.53 \\ \hline
%%%%%%%%%%%%%%%%%%%%%%%%%%%%%%%%%%%%%%%%%%%%%%%%%%%%
\multicolumn{5}{|l|}{\textbf{Sequential Models on \corpussmall train set}} \\ \hline
Sent2Vec + BiGRU+ att. & 60.05 & 55.8 & 57.85 & 55.67 \\ \hline
Doc2vec + BiGRU + att.  & 58.07 & 57.44 & 57.75 & 59.23 \\ \hline
\textbf{GloVe + BiGRU + att.} & 66.92 & 62.30 & \textbf{64.53} & 62.2 \\ \hline
HAN & 57.64 & 55.56 & 56.58 & 55.44 \\ \hline
\begin{tabular}[c]{@{}c@{}}Catchphrases + Sent2Vec\\  + BiGRU + att.\end{tabular}  & 61.90  & 60.13  & 61.00 & 60.06 \\ \hline
%HAN & X & X  & X & X & X \\ \hline
%%%%%%%%%%%%%%%%%%%%%%%%%%%%%%%%%%%%%%%%%%%%%%%%%%%%
\multicolumn{5}{|l|}{\textbf{Transformer Models on \corpuslarge train set}} \\ \hline
BERT Base & 60.56 & 57.64 & 59.06 & 57.65 \\ \hline
BERT Base & 67.54  & 62.22 & 64.77 & 62.10 \\ \hline
BERT Base  & 67.24& 63.85 & 65.50  & 63.74 \\ \hline
BERT Base  & 66.12  & 60.58 & 63.23  & 60.45 \\ \hline
BERT Base & 69.33 & 67.31  & 68.31& 67.24 \\ \hline
DistillBERT & 65.21 & 64.26  & 64.73 & 64.21  \\ \hline
\textbf{RoBERTa}  & 72.25 & 71.31 & \textbf{71.77} & 71.26 \\ \hline
XLNet  & 72.09  & 70.07 & 71.07 & 70.01 \\ \hline
%%%%%%%%%%%%%%%%%%%%%%%%%%%%%%%%%%%%%%%%%%%%%%%%%%%%
\multicolumn{5}{|l|}{\textbf{Hierarchical Models on \corpuslarge train set}}  \\ \hline
BERT + BiGRU  & 70.98 & 70.42 & 70.69 & 70.38 \\ \hline
RoBERTa + BiGRU & 75.13 & 74.30 & \begin{tabular}[c]{@{}c@{}}74.71 \\ ($\pm$0.01)\end{tabular} &
\begin{tabular}[c]{@{}c@{}}74.33 \\ ($\pm$1.99)\end{tabular}
\\ \hline
%74.71  ($\pm$0.01) & 74.33  ($\pm$1.99) \\ \hline
\textbf{XLNet + BiGRU}  & 77.80  & 77.78 & \textbf{77.79} & \textbf{77.78} \\ \hline
BERT + CNN & 71.68 & 70.17  & 70.92 & 70.12 \\ \hline
RoBERTa + CNN  & 74.74 & 73.17  & 73.95 & 73.22 \\ \hline
XLNet + CNN & 77.84& 77.21 & 77.53 & 77.24\\ \hline
%%%%%%%%%%%%%%%%%%%%%%%%%%%%%%%%%%%%%%%%%%%%%%%%%%%%
\multicolumn{5}{|l|}{\textbf{Hierarchical Models on \corpussmall train set}} \\ \hline
BERT + BiGRU & 65.28 & 63.95  & \begin{tabular}[c]{@{}c@{}}64.27\\($\pm$0.0116)\end{tabular}& \begin{tabular}[c]{@{}c@{}}63.89\\($\pm$1.10)\end{tabular} \\ \hline
RoBERTa + BiGRU & 73.24 & 72.93  & \begin{tabular}[c]{@{}c@{}}73.09\\($\pm$0.0022) \end{tabular} & \begin{tabular}[c]{@{}c@{}}72.95\\($\pm$0.25)\end{tabular} \\ \hline
XLNet + BiGRU & 75.11 & 75.06 & \begin{tabular}[c]{@{}c@{}}\textbf{75.09} \\ \textbf{($\pm$0.0043)} \end{tabular} & \begin{tabular}[c]{@{}c@{}}75.06\\($\pm$0.42)\end{tabular} \\ \hline
%%%%%%%%%%%%%%%%%%%%%%%%%%%%%%%%%%%%%%%%%%%%%%%%%%%%
\multicolumn{5}{|l|}{\textbf{Hierarchical Models with Attention on \corpuslarge train set}} \\ \hline
BERT + BiGRU + att. & 71.31 & 70.98 & \begin{tabular}[c]{@{}c@{}}71.14\\($\pm$0.0011)\end{tabular} & \begin{tabular}[c]{@{}c@{}}71.26\\($\pm$0.09) \end{tabular}\\ \hline
RoBERTa + BiGRU + att.  & 75.89 & 74.88 & \begin{tabular}[c]{@{}c@{}}75.38\\($\pm$0.0004) \end{tabular} & \begin{tabular}[c]{@{}c@{}}74.91\\($\pm$0.11)\end{tabular} \\ \hline
\textbf{XLNet + BiGRU + att.} & 77.32  & 76.82  & \begin{tabular}[c]{@{}c@{}}\textbf{77.07}\\ \textbf{($\pm$0.0077)}\end{tabular} & \begin{tabular}[c]{@{}c@{}}77.01\\ ($\pm$0.52)\end{tabular} \\ \hline
%%%%%%%%%%%%%%%%%%%%%%%%%%%%%%%%%%%%%%%%%%%%%%%%%%%%
\multicolumn{5}{|l|}{\textbf{Hierarchical Models with Attention on \corpussmall train set}}  \\ \hline
BERT + BiGRU + att. & 68.30 & 62.05  & \begin{tabular}[c]{@{}c@{}}65.03\\($\pm$0.0084)\end{tabular} & \begin{tabular}[c]{@{}c@{}}61.93\\($\pm$0.68)\end{tabular} \\ \hline
RoBERTa + BiGRU + att.  & 73.39 & 72.66 & \begin{tabular}[c]{@{}c@{}}73.02\\($\pm$0.0017)\end{tabular} & \begin{tabular}[c]{@{}c@{}}72.69\\($\pm$0.29)\end{tabular} \\ \hline
\textbf{XLNet + BiGRU + att.}    & 75.26 & 75.22 & \begin{tabular}[c]{@{}c@{}}\textbf{75.25}\\ \textbf{($\pm$0.0009)} \end{tabular} & \begin{tabular}[c]{@{}c@{}}75.22\\($\pm$0.13)\end{tabular} \\ \hline
%%%%%%%%%%%%%%%%%%%%%%%%%%%%%%%%%%%%%%%%%%%%%%%%%%%%
\multicolumn{5}{|l|}{\textbf{Transformers Voting Ensemble}} \\ \hline
\textbf{RoBERTa} & 68.20 & 62.55 & \textbf{65.26} & 62.43 \\ \hline
XLNet & 67.84  & 60.07 & 63.72 & 59.92  \\ \hline
%%%%%%%%%%%%%%%%%%%%%%%%%%%%%%%%%%%%%%%%%%%%%%%%%%%%
\multicolumn{5}{|l|}{\textbf{Hierarchical concatenated model with attention on \corpussmall train}} \\ \hline
\textbf{XLNet + BiGRU} & 76.85 & 76.31 & \begin{tabular}[c]{@{}c@{}}\textbf{76.55}\\ \textbf{($\pm$0.0140)}\end{tabular} & \begin{tabular}[c]{@{}c@{}}76.32\\($\pm$2.43)\end{tabular}\\ \hline
%%%%%%%%%%%%%%%%%%%%%%%%%%%%%%%%%%%%%%%%%%%%%%%%%%%%

\end{tabular}}
%\end{center}
% \caption{Prediction Results using different models.}
\caption{Prediction Results using different models. Some of the transformer and hierarchical models vary in performance across runs, we average out performance across 3 runs (variance in the parenthesis).} %\SN{We can remove hyper-parameter column and can make separate table for same at appendix. That will save one entire column.} \AB{agree --  we can move}}
\vspace{-3mm}
\label{tab:results}
\end{table}

\noindent\textbf{Sequence Models:} We experimented with standard BiGRU (2 layers) with attention model. We tried 3 different types of embeddings: (i)~Word level trained GloVe embeddings \cite{pennington2014glove}, with last 512 tokens as input, (ii)~Sentence level embeddings (Sent2Vec), where last 150 sentences were input\footnote{last 150 sentences covered around 90\% of the documents}, and (iii)~Chunk level embeddings (trained via Doc2Vec). We also trained Hierarchical Attention Network (HAN) \cite{yang2016hierarchical} model. GloVe embeddings with BiGRU and attention model gave the best performance (64\% F1) among the sequential models. Sequential models trained on \corpuslarge and \corpussmall have similar performances

\noindent \textbf{Transformer Models:} We experimented with BERT \cite{devlin-etal-2019-bert}, DistilBERT \cite{sanh2019distilbert}, RoBERTa \cite{liu2019roberta}, and XLNet \cite{yang2019xlnet}. Due to limitation on the number of input tokens to BERT and other transformer models, we experimented with different sections (begin tokens, middle tokens, end tokens, combinations of these) of the documents and as shown in Table \ref{tab:results}, the last 512 tokens gave the best performance. In general, transformer models outperform classical and sequential models. RoBERTa gave the best performance (72\% F1) and DistilBERT was the worst. We did not experiment with domain specific transformers like LEGAL-BERT \citep{chalkidis-etal-2020-legal}, since these have been trained upon US/EU legal texts, hence, they do not work well in the Indian setting as the legal systems are entirely different.

\begin{figure}
    \centering
    \includegraphics[scale=0.25]{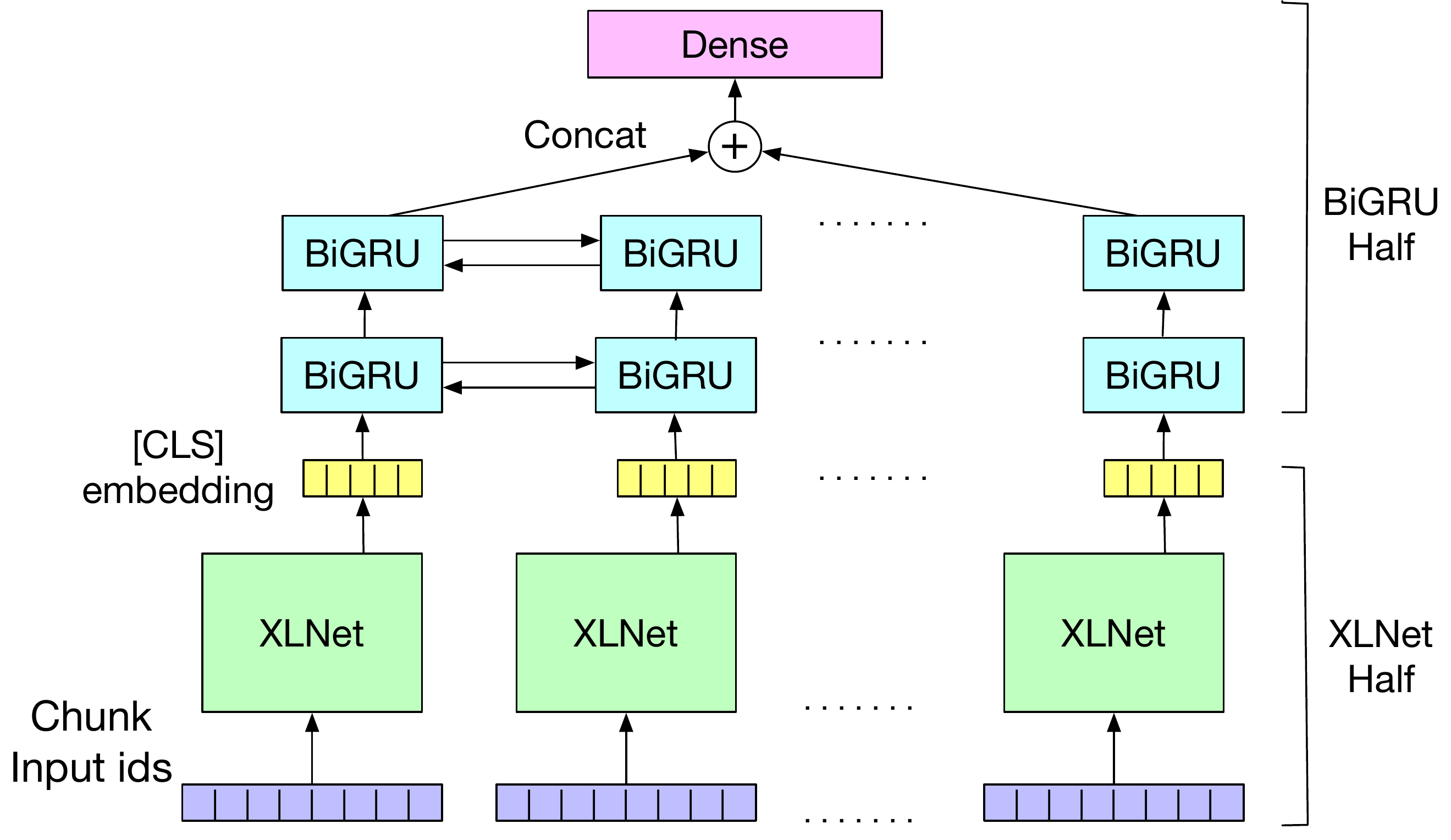}
    \vspace{-1mm}
    \caption{Hierarchical XLNet architecture (XLNet + BiGRU)} %\AB{Figure size is too small}
    \label{fig:xlnet}
    \vspace{-5mm}
\end{figure}

\noindent \textbf{Hierarchical Transformer Models:} Taking inspiration from hierarchical topic prediction model \cite{chitkara-etal-2019-topic}, we developed Hierarchical Transformer model architecture \cite{chalkidis-etal-2019-neural}. We divided each document into chunks using a moving window approach where each chunk was of length 512 tokens, and there was an overlap of 100 tokens. We obtained the $[CLS]$ representation of these chunks, which were then used as input to sequential models (BiGRU + attention) or feed-forward model (CNN \cite{kim-2014-convolutional}). We also tried an ensemble of individual transformer models on each of the chunks. 

In general, all the hierarchical models outperform transformer models. The best performing model (78\% F1) for predicting the case decision is XLNet with BiGRU on the top (Figure \ref{fig:xlnet}). Comparing best model accuracy with average annotator accuracy (78\% vs. 94\%) indicates the task's inherent complexity and motivates more research in this direction.  

% \vspace{-3.5mm}
\subsection{Case Decision Explanation} \label{sec:explanation-model}

\noindent We experimented with a variety explainability algorithms as a post-prediction step. We experimented with the best judgment prediction model (Hierarchical Transformer (XLNet + BiGRU)) for all the explainable algorithms. We explored three class of explainability methods \citep{xie2020explainable}: attribution based, model agnostic, and attention-based. 

In the class of attribution based methods, Layerwise Relevance Propagation (LRP) \cite{bach2015pixel} and DeepLIFT \cite{shrikumar2017learning} methods did not work in our case. Due to the long length of documents, model agnostic explainability methods like LIME \cite{ribeiro2016should} and Anchors \cite{DBLP:conf/aaai/Ribeiro0G18} were not applicable. We also experimented with attention-based methods, and Integrated Gradients \cite{sundararajan2017axiomatic} method using the CAPTUM library \cite{captum2019github}. %\RS{As we have removed the reference of CAPTUM from appendix so we need to edit here}  
However, these highlighted only a few tokens or short phrases. %(see \cref{supp:explain}).
Moreover, attention-based scores are not necessarily indicative of explanations \cite{jain-wallace-2019-attention}.  

To extract explanations, we propose a method inspired from  \newcite{li2016understanding} and \newcite{zeiler2014visualizing}. The idea is to use the occlusion method at both levels of the hierarchy. For each document, for the BiGRU part of the model, we mask each complete chunk embedding one at a time. The masked input is passed through the trained BiGRU, and the output probability (masked probability) of the label obtained by the original unmasked model is calculated. The masked probability is compared with unmasked probability to calculate the chunk explainability score. Formally, for a chunk $c$, if the sigmoid outputs (of the BiGRU) are $\sigma_{m}$ (when the chunk was not masked) and $\sigma_{m'}$ (when the chunk was masked) and the predicted label is ${y}$  then the probabilities and chunk score
$s_{c} = p_{m} - p_{m'}$ and $p_{m'/m} =\begin{cases}
			\sigma_{m'/m}, & \hspace{-2mm} y = 1\\
            1 - \sigma_{m'/m}, & \hspace{-2mm} y = 0
		 \end{cases}$

We obtain sentences that explain the decision from the transformer part of the model (XLNet) using the chunks that were assigned positive scores. Each chunk (length 512 tokens) is segmented into sentences using NLTK sentence splitter \cite{loper2002nltk}. Similar to BiGRU, each sentence is masked and the output of the transformer at the classification head (softmax logits) is compared with logits of the label corresponding to original hierarchical model. The difference between the logits normalized by the length of the sentence is the explanation score of the sentence. Finally, top-k sentences ($\sim$ 40\%) in each chunk are selected.
% The same process is repeated for the XLNet, where for each chunk, a sentence is occluded, and a normalized difference in logits is used as the explanation score.  
% \begin{figure}[t]%[htbp]
% \vspace{-3mm}
%      \begin{subfigure}[b]{0.24\textwidth}
%          \centering
%          \includegraphics[width=\textwidth]{images/occ10chunk.png}
%          \vspace{-7mm}
%          \subcaption{Occlusion}
%          \vspace{-4mm}
%      \end{subfigure}\hfill
%      \begin{subfigure}[b]{0.24\textwidth}
%          \centering
%          \includegraphics[width=\textwidth]{images/att10chunk.png}
%           \vspace{-7mm}
%          \subcaption{Attention}
%          \vspace{-4mm}
%      \end{subfigure}
%         \caption{Averaged chunk scores}
%         \label{fig:attention_occlusion_graphs}
%         \vspace{-4mm}
% \end{figure}

% \usepackage{multirow}
% \usepackage{multirow}
% \usepackage{multirow}
\begin{table}[t]
\small
\begin{tabular}{|c|c|c|c|c|c|}
\hline
	\multirow{3}{*}{\textbf{Metric}}                                      & \multicolumn{5}{c|}{\textbf{Explainability Model vs Experts}}    \\ \cline{2-6} 
                                                                      & \multicolumn{5}{c|}{\textbf{Expert}}                             \\ \cline{2-6} 
                                                                      & \textbf{1} & \textbf{2} & \textbf{3} & \textbf{4} & \textbf{5} \\ \hline
\textbf{\begin{tabular}[c]{@{}c@{}}Jaccard\\ Similarity\end{tabular}} & 0.333      & 0.317      & 0.328      & 0.324      & 0.318      \\ \hline
\textbf{Overlap-Min}                                                  & 0.744      & 0.589      & 0.81       & 0.834      & 0.617      \\ \hline
\textbf{Overlap-Max}                                                  & 0.39       & 0.414      & 0.36       & 0.35       & 0.401      \\ \hline
\textbf{ROUGE-1}                                                      & 0.444      & 0.517      & 0.401      & 0.391      & 0.501      \\ \hline
\textbf{ROUGE-2}                                                      & 0.303      & 0.295      & 0.296      & 0.297      & 0.294      \\ \hline
\textbf{ROUGE-L}                                                      & 0.439      & 0.407      & 0.423      & 0.444      & 0.407      \\ \hline
\textbf{BLEU}                                                         & 0.16       & 0.28       & 0.099      & 0.093      & 0.248      \\ \hline
\textbf{Meteor}                                                       & 0.22       & 0.3        & 0.18       & 0.177      & 0.279      \\ \hline
\end{tabular}
\vspace{-3mm}
\caption{Machine explanations v/s Expert explanations}
\label{tab:explainabilityVsAnnotators}
\vspace{-5mm}
\end{table}

\begin{figure}[t]%[htbp]
%\vspace{-3mm}
         \centering
         \includegraphics[scale=0.25]{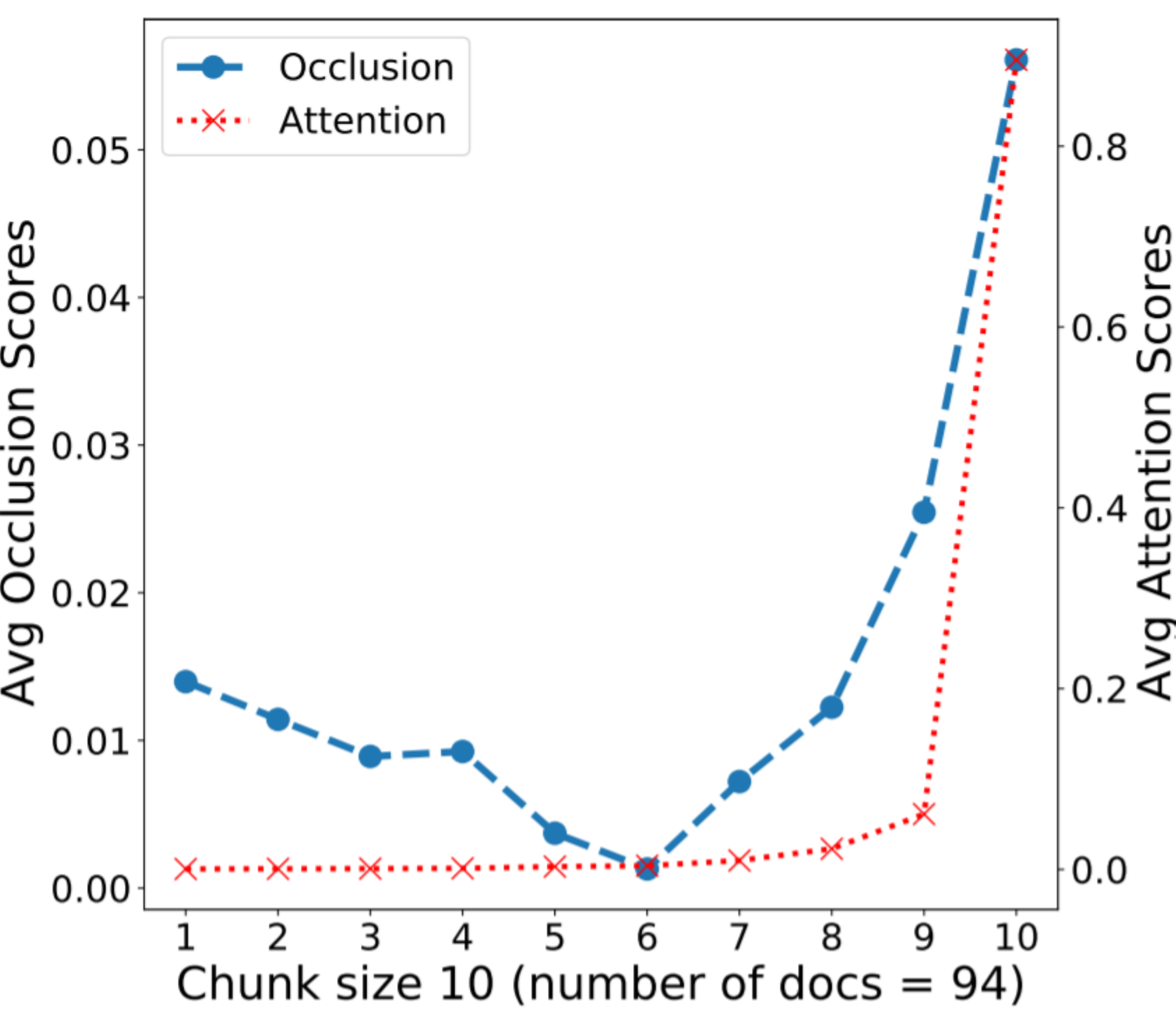}         %\vspace{-3mm}
        \caption{Averaged chunk scores for attention and occlusion}
        \label{fig:attention_occlusion_graphs}
        \vspace{-5mm}
\end{figure}

To understand and analyze which parts of the documents were contributing towards prediction, we examined the attention weights (scores) in the case of the XLNet+BiGRU+Attention model and the occlusion scores of the XLNet+BiGRU model. Plots for some of the documents are shown in Figure \ref{fig:attention_occlusion_graphs}. Plots for different chunk sizes are provided in Data/images folder in our GitHub repository. We also provide the t-SNE visualization on the test set using the BERT and Doc2Vec embeddings. Token visualization heatmap using Integrated Gradient for document name 1951\_33.txt for BERT model is also provided in GitHub. Plots of scores averaged out over the entire test set for each chunk size can be visualized in \cref{supp:explain}. Two things can be noted: firstly, the largest attention and occlusion scores are assigned to chunks corresponding to the end of the document; this is in line with our hypothesis that most of the important information and rationale for judgment is mainly towards the end of the document. Secondly, although attention scores are optimized (via loss minimization or accuracy maximization) to concentrate on the last chunks, this is not the case with occlusion scores. There is no optimization of occlusion scores; yet they still focus on the chunks at the end, which affirms our hypothesis. %One might argue that this observation might be due to the transformer being trained on last 512 tokens only. To check this, we also visualized the hierarchical transformers trained on \corpussmall (see \cref{sup:hmodel}), but the results were similar as to what we have observed in this case.  

%\vspace{-3.2mm}
\subsection{Model Explainability versus Annotators}\label{sec:explainabilitymodelVsAnnotators}
We compare the performance of occlusion method explanations with the expert annotators' gold explanations by measuring the overlap between the two. We used the same measures (\cref{sec:analysis-anno}) ROUGE-L, ROUGE-1, ROUGE-2, Jaccard Similarity, 
BLEU, METEOR, Overlap Maximum, and Overlap Minimum %\RS{Sir, here also do we need to remove some of the heatmaps?}. 
Table \ref{tab:explainabilityVsAnnotators} compares machine explanations with the gold explanations. The highest overlap value ($0.8337$) is observed for the measure Overlap-Min with Expert 4. The values for Overlap-Min depict high agreements of the explainability model with all the experts. However, the values for the other evaluation measures, e.g., ROUGE-L, are in the low to medium range, the highest being $0.4445$ for ROUGE-L and Expert 4. The results show the wide gap between how a machine would explain a judgment and the way a legal expert would explain it. The results motivate us for future research in this direction of developing an explainable model. 

% \begin{figure}[t]%[htbp]
% %\vspace{-3mm}
%      \begin{subfigure}[b]{0.24\textwidth}
%          \centering
%          \includegraphics[width=\textwidth]{images/occ_10th_cropped.pdf}
%          \vspace{-7mm}
%          \subcaption{Occlusion}
%          \vspace{-4mm}
%      \end{subfigure}\hfill
%      \begin{subfigure}[b]{0.24\textwidth}
%          \centering
%          \includegraphics[width=\textwidth]{images/att_10_cropped.pdf}
%           \vspace{-7mm}
%          \subcaption{Attention}
%          \vspace{-4mm}
%      \end{subfigure}
%         \caption{Averaged chunk scores}
%         \label{fig:attention_occlusion_graphs}
%         \vspace{-4mm}
% \end{figure}

\vspace{-3mm}

%\section{Conclusion and Future Directions}
\section{Conclusion}
\vspace{-2mm}
This paper introduces the ILDC corpus and corresponding CJPE task. The corpus is annotated with case decisions and explanations for the decisions for a separate test set. Analysis of the corpus and modeling results shows the complexity of legal documents that pose challenges from a computational perspective. We hope that the corpus and the task would provide a challenging and interesting resource for the Legal NLP researchers. For future work, we would like to train a legal transformer similar to LEGAL-BERT \citep{chalkidis-etal-2020-legal} on our Indian legal case documents. Moreover, we would also like to focus upon using rhetorical roles \newcite{bhattacharya2019identification} of the sentences to include structural information of the documents for CJPE task as well. %In the current work, we just used data-driven approaches for attempting prediction and explanation tasks; however, in the future, we would like to explore how prior knowledge in the form of written laws/legal-articles could be integrated. Another interesting direction to explore would be a hybrid system based on neuro-symbolic reasoning \citep{DBLP:journals/corr/abs-1711-03902}.   

%This raises interesting research questions about developing explanation methods which align with the experts' way of thinking. \AB{don't we need a future work kind of statement here?}
%This possibly leaves scope for future work in automating explanations for the \taskname task.
 %In future, we would like to expand the dataset. Also, we would like to explore if other explainability models can be applied, and the prediction task can be improved by employing semantic segmentation of the case documents (e.g., detecting rhetorical roles).
\vspace{-2mm}

\section*{Acknowledgements}
\vspace{-2mm}
We would like to thank anonymous reviewers for their insightful comments. We would like to thank student research assistants Abin Thomas Alex, Amrita Ghosh, Parmeet Singh, and Unnati Jhunjhunwala from West Bengal National University of Juridical Sciences (WBNUJS) for annotating the documents. This work would not have been possible without their help. 

\section*{Ethical Concerns} \label{sec:ethics}

%The system intends to assist legal professionals in their research and decision-making and not replace them. Expert human analysis of the output would ensure an additional layer of ethical scrutiny. Bias safeguards employed (described in \cref{sec:ildc}) also partly address ethical concerns. To the best of our knowledge, such ethical concerns related to CJPE (and LJP in general) is still a new and unexplored area. Moreover, ethical consideration of legal case prediction are a research area on its own (and is not limited to just NLP domain) and the community needs to explore more on this. The proposed corpus is an initial attempt towards this research. 
The corpus is created from publicly available data: proceedings of Supreme Court of India (SCI). The data was scraped from the website: \url{www.indiankanoon.org}. The website allows scrapping of the data and no copyrights were infringed. Annotators were selected randomly and they participated voluntarily. 

The proposed corpus aims to promote the development of an explainable case judgment prediction system. The system intends to assist legal professionals in their research and decision-making and not replace them. Therefore, ethical considerations such as allowing legal rights and obligations of human beings to be decided and pronounced upon by non-human intelligence are not being breached by the system. The system proposes to provide valuable information that might be useful to a legal professional to make strategic decisions, but the actual decision-making process is still going to be carried out by the professional himself. Therefore, the system is not intended to produce a host of artificial lawyers and judges regulating human behavior. At the same time, the final expert human analysis of the systemic output should ensure that any existing flaw, absurdity, or overt or latent bias gets subjected to an additional layer of ethical scrutiny. In this way, the usual ethical concerns associated with the concept of case-law prediction also get addressed to a considerable extent since the system is not performing any judicial role herein nor deciding the legal rights or liabilities of human beings. Instead, the system is purported to be used primarily by legal professionals to make strategic decisions of their own, said decisions being still subjected to legal and judicial scrutiny performed by human experts. Nevertheless, the community needs to pursue more research in this regard to fully understand the unforeseen social implications of such system. This paper takes initial steps by introducing the corpus and baseline models to the community.   

Care has been taken to select cases in a completely random manner, without any particular focus on the type of law or the identities or socio-politico-economic background of the parties or the judges involved. Specifically, the aforementioned identities have been deliberately anonymized so as to minimize or eliminate any possible bias in the course of prediction. The subjectivity that is associated with the judicial decision-making may also be controlled in this way, since the system is focusing on how consideration of the facts and applicable law are supposed to determine the outcome of the cases, instead of any individual bias on the judge’s part; another judge might not share such bias, and therefore the only common point of reference that the two judges would have would be the relevant facts of the case and the laws involved. This also gets reflected in the objective methodology used in the selection of annotators and by eliminating any interaction between the annotators themselves while at the same time paying attention to the factors or observations common to the output from the various annotators.

The only specification with regard to the forum has been made by taking all the cases from the domain of the Supreme Court of India, owing to the propensity of the apex court of the land towards focusing on the legalities of the issues involved rather than rendering mere fact-specific judgments, as well as the binding nature of such decisions on the subordinate courts of the land. This would also allow the results to be further generalized and applied to a broader set of cases filed before other forums, too, since the subordinate courts are supposed to follow the reasoning of the Supreme Court’s judgments to the greatest possible extent. As a result, the impact of the training and testing opportunities provided to the system by a few Supreme Court cases is likely to be much greater than the mere absolute numbers would otherwise suggest.

% \input{corpus}
% \vspace{-3mm}

% \vspace{2mm}

% \input{models}
% \vspace{-1.5mm}

% \input{results}
% \vspace{-2mm}
% % %\input{analysis}

% \input{future}

\bibliographystyle{acl_natbib}
%\bibliography{anthology,acl2021}
\bibliography{acl2021}

\begin{thebibliography}{56}
\expandafter\ifx\csname natexlab\endcsname\relax\def\natexlab#1{#1}\fi

\bibitem[{Aletras et~al.(2016)Aletras, Tsarapatsanis, Preotiuc{-}Pietro, and
  Lampos}]{Aletras2016}
Nikolaos Aletras, Dimitrios Tsarapatsanis, Daniel Preotiuc{-}Pietro, and
  Vasileios Lampos. 2016.
\newblock {P}redicting judicial decisions of the {E}uropean {C}ourt of {H}uman
  {R}ights:{{A} {N}atural {L}anguage {P}rocessing} perspective.
\newblock \emph{PeerJ Computer Science}, 2:93.

\bibitem[{Bach et~al.(2015)Bach, Binder, Montavon, Klauschen, M{\"u}ller, and
  Samek}]{bach2015pixel}
Sebastian Bach, Alexander Binder, Gr{\'e}goire Montavon, Frederick Klauschen,
  Klaus-Robert M{\"u}ller, and Wojciech Samek. 2015.
\newblock On pixel-wise explanations for non-linear classifier decisions by
  layer-wise relevance propagation.
\newblock \emph{PloS one}, 10(7):e0130140.

\bibitem[{Beltagy et~al.(2020)Beltagy, Peters, and
  Cohan}]{beltagy2020longformer}
Iz~Beltagy, Matthew~E Peters, and Arman Cohan. 2020.
\newblock {L}ongformer: The long-document transformer.
\newblock \emph{arXiv preprint arXiv:2004.05150}.

\bibitem[{Bhattacharya et~al.(2019{\natexlab{a}})Bhattacharya, Hiware,
  Rajgaria, Pochhi, Ghosh, and Ghosh}]{bhattacharya2019comparative}
Paheli Bhattacharya, Kaustubh Hiware, Subham Rajgaria, Nilay Pochhi,
  Kripabandhu Ghosh, and Saptarshi Ghosh. 2019{\natexlab{a}}.
\newblock A comparative study of summarization algorithms applied to legal case
  judgments.
\newblock In \emph{European Conference on Information Retrieval}, pages
  413--428. Springer.

\bibitem[{Bhattacharya et~al.(2019{\natexlab{b}})Bhattacharya, Paul, Ghosh,
  Ghosh, and Wyner}]{bhattacharya2019identification}
Paheli Bhattacharya, Shounak Paul, Kripabandhu Ghosh, Saptarshi Ghosh, and Adam
  Wyner. 2019{\natexlab{b}}.
\newblock {I}dentification of {R}hetorical {R}oles of {S}entences in {I}ndian
  {L}egal {J}udgments.
\newblock In \emph{Legal Knowledge and Information Systems - {JURIX} 2019},
  volume 322 of \emph{Frontiers in Artificial Intelligence and Applications},
  pages 3--12. {IOS} Press.

\bibitem[{de~Castilho et~al.(2016)de~Castilho, Mujdricza-Maydt, Yimam,
  Hartmann, Gurevych, Frank, and Biemann}]{de2016web}
Richard~Eckart de~Castilho, Eva Mujdricza-Maydt, Seid~Muhie Yimam, Silvana
  Hartmann, Iryna Gurevych, Anette Frank, and Chris Biemann. 2016.
\newblock A web-based tool for the integrated annotation of semantic and
  syntactic structures.
\newblock In \emph{Proceedings of the Workshop on Language Technology Resources
  and Tools for Digital Humanities (LT4DH)}, pages 76--84.

\bibitem[{Chalkidis et~al.(2019)Chalkidis, Androutsopoulos, and
  Aletras}]{chalkidis-etal-2019-neural}
Ilias Chalkidis, Ion Androutsopoulos, and Nikolaos Aletras. 2019.
\newblock {N}eural {L}egal {J}udgment {P}rediction in {E}nglish.
\newblock In \emph{Proceedings of the 57th Annual Meeting of the Association
  for Computational Linguistics}, pages 4317--4323, Florence, Italy.
  Association for Computational Linguistics.

\bibitem[{Chalkidis et~al.(2020)Chalkidis, Fergadiotis, Malakasiotis, Aletras,
  and Androutsopoulos}]{chalkidis-etal-2020-legal}
Ilias Chalkidis, Manos Fergadiotis, Prodromos Malakasiotis, Nikolaos Aletras,
  and Ion Androutsopoulos. 2020.
\newblock {LEGAL}-{BERT}: The {M}uppets straight out of {L}aw {S}chool.
\newblock In \emph{Findings of the Association for Computational Linguistics:
  EMNLP 2020}, pages 2898--2904, Online. Association for Computational
  Linguistics.

\bibitem[{Chen et~al.(2019)Chen, Cai, Dai, Dai, and
  Ding}]{chen-etal-2019-charge}
Huajie Chen, Deng Cai, Wei Dai, Zehui Dai, and Yadong Ding. 2019.
\newblock {C}harge-{B}ased {P}rison {T}erm {P}rediction with {D}eep {G}ating
  {N}etwork.
\newblock In \emph{Proceedings of the 2019 Conference on Empirical Methods in
  Natural Language Processing and the 9th International Joint Conference on
  Natural Language Processing (EMNLP-IJCNLP)}, pages 6362--6367, Hong Kong,
  China. Association for Computational Linguistics.

\bibitem[{Chitkara et~al.(2019)Chitkara, Modi, Avvaru, Janghorbani, and
  Kapadia}]{chitkara-etal-2019-topic}
Pooja Chitkara, Ashutosh Modi, Pravalika Avvaru, Sepehr Janghorbani, and
  Mubbasir Kapadia. 2019.
\newblock {T}opic {S}potting using {H}ierarchical {N}etworks with {S}elf
  {A}ttention.
\newblock In \emph{Proceedings of the 2019 Conference of the North {A}merican
  Chapter of the Association for Computational Linguistics: Human Language
  Technologies, Volume 1 (Long and Short Papers)}, pages 3755--3761,
  Minneapolis, Minnesota. Association for Computational Linguistics.

\bibitem[{Devlin et~al.(2019)Devlin, Chang, Lee, and
  Toutanova}]{devlin-etal-2019-bert}
Jacob Devlin, Ming-Wei Chang, Kenton Lee, and Kristina Toutanova. 2019.
\newblock {BERT}: {P}re-training of {D}eep {B}idirectional {T}ransformers for
  {L}anguage {U}nderstanding.
\newblock In \emph{Proceedings of the 2019 Conference of the North {A}merican
  Chapter of the Association for Computational Linguistics: Human Language
  Technologies, Volume 1 (Long and Short Papers)}, pages 4171--4186,
  Minneapolis, Minnesota. Association for Computational Linguistics.

\bibitem[{Fleiss(1971)}]{FleissKappa}
J.~L. Fleiss. 1971.
\newblock {M}easuring nominal scale agreement among many raters.
\newblock \emph{Psychological Bulletin}, 75(5).

\bibitem[{Galgani et~al.(2012)Galgani, Compton, and
  Hoffmann}]{galgani2012towards}
Filippo Galgani, Paul Compton, and Achim Hoffmann. 2012.
\newblock {T}owards automatic generation of catchphrases for legal case
  reports.
\newblock In \emph{International Conference on Intelligent Text Processing and
  Computational Linguistics}, pages 414--425. Springer.

\bibitem[{Hu et~al.(2018)Hu, Li, Tu, Liu, and Sun}]{ChargePredictionFewShot}
Zikun Hu, Xiang Li, Cunchao Tu, Zhiyuan Liu, and Maosong Sun. 2018.
\newblock {F}ew-{S}hot {C}harge {P}rediction with {D}iscriminative {L}egal
  {A}ttributes.
\newblock In \emph{Proceedings of the 27th International Conference on
  Computational Linguistics}, pages 487--498, Santa Fe, New Mexico, USA.
  Association for Computational Linguistics.

\bibitem[{Jackson et~al.(2003)Jackson, Al-Kofahi, Tyrrell, and
  Vachher}]{jackson2003information}
Peter Jackson, Khalid Al-Kofahi, Alex Tyrrell, and Arun Vachher. 2003.
\newblock {I}nformation extraction from case law and retrieval of prior cases.
\newblock \emph{Artificial Intelligence}, 150(1-2):239--290.

\bibitem[{Jain and Wallace(2019)}]{jain-wallace-2019-attention}
Sarthak Jain and Byron~C. Wallace. 2019.
\newblock {A}ttention is not {E}xplanation.
\newblock In \emph{Proceedings of the 2019 Conference of the North {A}merican
  Chapter of the Association for Computational Linguistics: Human Language
  Technologies, Volume 1 (Long and Short Papers)}, pages 3543--3556,
  Minneapolis, Minnesota. Association for Computational Linguistics.

\bibitem[{Jiang et~al.(2018)Jiang, Ye, Luo, Chao, and
  Ma}]{jiang-interpretableChargePrediction}
Xin Jiang, Hai Ye, Zhunchen Luo, WenHan Chao, and Wenjia Ma. 2018.
\newblock {I}nterpretable {R}ationale {A}ugmented {C}harge {P}rediction
  {S}ystem.
\newblock In \emph{Proceedings of the 27th International Conference on
  Computational Linguistics: System Demonstrations}, pages 146--151, Santa Fe,
  New Mexico. Association for Computational Linguistics.

\bibitem[{Katju(2019)}]{backlog-cases2019}
Justice~Markandey Katju. 2019.
\newblock Backlog of cases crippling judiciary.
\newblock \url{https://tinyurl.com/v4xu6mvk}.

\bibitem[{Katz et~al.(2017)Katz, Bommarito, and Blackman}]{Katz2017}
Daniel~Martin Katz, Michael~J. Bommarito, II, and Josh Blackman. 2017.
\newblock {A} general approach for predicting the behavior of the {S}upreme
  {C}ourt of the {U}nited {S}tates.
\newblock \emph{PLOS ONE}, 12:1--18.

\bibitem[{Kim(2014)}]{kim-2014-convolutional}
Yoon Kim. 2014.
\newblock {C}onvolutional {N}eural {N}etworks for {S}entence {C}lassification.
\newblock In \emph{Proceedings of the 2014 Conference on Empirical Methods in
  Natural Language Processing ({EMNLP})}, pages 1746--1751, Doha, Qatar.
  Association for Computational Linguistics.

\bibitem[{Kitaev et~al.(2020)Kitaev, Kaiser, and
  Levskaya}]{Kitaev2020Reformer:}
Nikita Kitaev, Lukasz Kaiser, and Anselm Levskaya. 2020.
\newblock Reformer: {T}he {E}fficient {T}ransformer.
\newblock In \emph{International Conference on Learning Representations}.

\bibitem[{Kokhlikyan et~al.(2019)Kokhlikyan, Miglani, Martin, Wang, Reynolds,
  Melnikov, Lunova, and Reblitz-Richardson}]{captum2019github}
Narine Kokhlikyan, Vivek Miglani, Miguel Martin, Edward Wang, Jonathan
  Reynolds, Alexander Melnikov, Natalia Lunova, and Orion Reblitz-Richardson.
  2019.
\newblock {P}ytorch {C}aptum.
\newblock \url{https://github.com/pytorch/captum}.

\bibitem[{Lavie and Agarwal(2007)}]{lavie2007meteor}
Alon Lavie and Abhaya Agarwal. 2007.
\newblock {METEOR}: An automatic metric for {MT} evaluation with high levels of
  correlation with human judgments.
\newblock In \emph{Proceedings of the second workshop on statistical machine
  translation}, pages 228--231.

\bibitem[{Le and Mikolov(2014)}]{le2014distributed}
Quoc Le and Tomas Mikolov. 2014.
\newblock Distributed representations of sentences and documents.
\newblock In \emph{International conference on machine learning}, pages
  1188--1196.

\bibitem[{Li et~al.(2016)Li, Monroe, and Jurafsky}]{li2016understanding}
Jiwei Li, Will Monroe, and Dan Jurafsky. 2016.
\newblock Understanding neural networks through representation erasure.
\newblock \emph{arXiv preprint arXiv:1612.08220}.

\bibitem[{Lin(2004)}]{lin2004rouge}
Chin-Yew Lin. 2004.
\newblock {ROUGE}: A package for automatic evaluation of summaries.
\newblock In \emph{Text summarization branches out}, pages 74--81.

\bibitem[{Liu et~al.(2019)Liu, Ott, Goyal, Du, Joshi, Chen, Levy, Lewis,
  Zettlemoyer, and Stoyanov}]{liu2019roberta}
Yinhan Liu, Myle Ott, Naman Goyal, Jingfei Du, Mandar Joshi, Danqi Chen, Omer
  Levy, Mike Lewis, Luke Zettlemoyer, and Veselin Stoyanov. 2019.
\newblock Ro{BERT}a: {A} robustly optimized {BERT} pretraining approach.
\newblock \emph{arXiv preprint arXiv:1907.11692}.

\bibitem[{Long et~al.(2019)Long, Tu, Liu, and Sun}]{long2019automatic}
Shangbang Long, Cunchao Tu, Zhiyuan Liu, and Maosong Sun. 2019.
\newblock {A}utomatic judgment prediction via legal reading comprehension.
\newblock In \emph{China National Conference on Chinese Computational
  Linguistics}, pages 558--572. Springer.

\bibitem[{Loper and Bird(2002)}]{loper2002nltk}
Edward Loper and Steven Bird. 2002.
\newblock {NLTK}: the natural language toolkit.
\newblock \emph{arXiv preprint cs/0205028}.

\bibitem[{Luo et~al.(2017)Luo, Feng, Xu, Zhang, and
  Zhao}]{ChargePredictionLegalBasis}
Bingfeng Luo, Yansong Feng, Jianbo Xu, Xiang Zhang, and Dongyan Zhao. 2017.
\newblock {L}earning to {P}redict {C}harges for {C}riminal {C}ases with {L}egal
  {B}asis.
\newblock In \emph{Proceedings of the 2017 Conference on Empirical Methods in
  Natural Language Processing}, pages 2727--2736, Copenhagen, Denmark.
  Association for Computational Linguistics.

\bibitem[{Mandal et~al.(2017)Mandal, Ghosh, Pal, and
  Ghosh}]{mandal2017automatic}
Arpan Mandal, Kripabandhu Ghosh, Arindam Pal, and Saptarshi Ghosh. 2017.
\newblock Automatic catchphrase identification from legal court case documents.
\newblock In \emph{Proceedings of the 2017 ACM on Conference on Information and
  Knowledge Management}, pages 2187--2190.

\bibitem[{Pagliardini et~al.(2018)Pagliardini, Gupta, and
  Jaggi}]{pagliardini-etal-2018-unsupervised}
Matteo Pagliardini, Prakhar Gupta, and Martin Jaggi. 2018.
\newblock Unsupervised {L}earning of {S}entence {E}mbeddings {U}sing
  {C}ompositional n-{G}ram features.
\newblock In \emph{Proceedings of the 2018 Conference of the North {A}merican
  Chapter of the Association for Computational Linguistics: Human Language
  Technologies, Volume 1 (Long Papers)}, pages 528--540, New Orleans,
  Louisiana. Association for Computational Linguistics.

\bibitem[{Papineni et~al.(2002)Papineni, Roukos, Ward, and
  Zhu}]{papineni2002bleu}
Kishore Papineni, Salim Roukos, Todd Ward, and Wei-Jing Zhu. 2002.
\newblock {BLEU}: a method for automatic evaluation of machine translation.
\newblock In \emph{Proceedings of the 40th annual meeting of the Association
  for Computational Linguistics}, pages 311--318.

\bibitem[{Pedregosa et~al.(2011)Pedregosa, Varoquaux, Gramfort, Michel,
  Thirion, Grisel, Blondel, Prettenhofer, Weiss, Dubourg, Vanderplas, Passos,
  Cournapeau, Brucher, Perrot, and Duchesnay}]{scikit-learn}
F.~Pedregosa, G.~Varoquaux, A.~Gramfort, V.~Michel, B.~Thirion, O.~Grisel,
  M.~Blondel, P.~Prettenhofer, R.~Weiss, V.~Dubourg, J.~Vanderplas, A.~Passos,
  D.~Cournapeau, M.~Brucher, M.~Perrot, and E.~Duchesnay. 2011.
\newblock Scikit-learn: {M}achine {L}earning in {P}ython.
\newblock \emph{Journal of Machine Learning Research}, 12:2825--2830.

\bibitem[{Pennington et~al.(2014)Pennington, Socher, and
  Manning}]{pennington2014glove}
Jeffrey Pennington, Richard Socher, and Christopher~D Manning. 2014.
\newblock {G}lo{V}e: {G}lobal vectors for word representation.
\newblock In \emph{Proceedings of the 2014 conference on empirical methods in
  natural language processing (EMNLP)}, pages 1532--1543.

\bibitem[{Ribeiro et~al.(2016)Ribeiro, Singh, and Guestrin}]{ribeiro2016should}
Marco~Tulio Ribeiro, Sameer Singh, and Carlos Guestrin. 2016.
\newblock "{W}hy should {I} trust you?" {E}xplaining the predictions of any
  classifier.
\newblock In \emph{Proceedings of the 22nd ACM SIGKDD international conference
  on knowledge discovery and data mining}, pages 1135--1144.

\bibitem[{Ribeiro et~al.(2018)Ribeiro, Singh, and
  Guestrin}]{DBLP:conf/aaai/Ribeiro0G18}
Marco~T{\'{u}}lio Ribeiro, Sameer Singh, and Carlos Guestrin. 2018.
\newblock {A}nchors: {H}igh-{P}recision {M}odel-{A}gnostic {E}xplanations.
\newblock In \emph{Proceedings of the Thirty-Second {AAAI} Conference on
  Artificial Intelligence, (AAAI-18)}, pages 1527--1535. {AAAI} Press.

\bibitem[{Sanh et~al.(2019)Sanh, Debut, Chaumond, and
  Wolf}]{sanh2019distilbert}
Victor Sanh, Lysandre Debut, Julien Chaumond, and Thomas Wolf. 2019.
\newblock Distil{BERT}, a distilled version of {BERT}: smaller, faster, cheaper
  and lighter.
\newblock \emph{arXiv preprint arXiv:1910.01108}.

\bibitem[{Shrikumar et~al.(2017)Shrikumar, Greenside, and
  Kundaje}]{shrikumar2017learning}
Avanti Shrikumar, Peyton Greenside, and Anshul Kundaje. 2017.
\newblock Learning important features through propagating activation
  differences.
\newblock In \emph{International Conference on Machine Learning}, pages
  3145--3153. PMLR.

\bibitem[{Strickson and de~la Iglesia(2020)}]{Strickson2020}
Benjamin Strickson and Beatriz de~la Iglesia. 2020.
\newblock {L}egal {J}udgement {P}rediction for {UK} {C}ourts.
\newblock \emph{{ICISS} 2020: The 3rd International Conference on Information
  Science and System, Cambridge, UK, March 19-22, 2020}, pages 204--209.

\bibitem[{{\c{S}}ulea et~al.(2017){\c{S}}ulea, Zampieri, Vela, and van
  Genabith}]{sulea2017}
Octavia-Maria {\c{S}}ulea, Marcos Zampieri, Mihaela Vela, and Josef van
  Genabith. 2017.
\newblock {P}redicting the {L}aw {A}rea and {D}ecisions of {F}rench {S}upreme
  {C}ourt {C}ases.
\newblock In \emph{Proceedings of the International Conference Recent Advances
  in Natural Language Processing, {RANLP} 2017}, pages 716--722, Varna,
  Bulgaria. INCOMA Ltd.

\bibitem[{Sundararajan et~al.(2017)Sundararajan, Taly, and
  Yan}]{sundararajan2017axiomatic}
Mukund Sundararajan, Ankur Taly, and Qiqi Yan. 2017.
\newblock Axiomatic attribution for deep networks.
\newblock In \emph{International Conference on Machine Learning}, pages
  3319--3328. PMLR.

\bibitem[{Tran et~al.(2019)Tran, Nguyen, and Satoh}]{Tran2019}
Vu~Tran, Minh~Le Nguyen, and Ken Satoh. 2019.
\newblock {B}uilding {L}egal {C}ase {R}etrieval {S}ystems with {L}exical
  {M}atching and {S}ummarization {U}sing {A} {P}re-{T}rained {P}hrase {S}coring
  {M}odel.
\newblock In \emph{Proceedings of the Seventeenth International Conference on
  Artificial Intelligence and Law}, ICAIL ’19, page 275–282, New York, NY,
  USA. Association for Computing Machinery.

\bibitem[{Vaswani et~al.(2017)Vaswani, Shazeer, Parmar, Uszkoreit, Jones,
  Gomez, Kaiser, and Polosukhin}]{vaswani2017attention}
Ashish Vaswani, Noam Shazeer, Niki Parmar, Jakob Uszkoreit, Llion Jones,
  Aidan~N Gomez, {\L}ukasz Kaiser, and Illia Polosukhin. 2017.
\newblock {A}ttention is all you need.
\newblock In \emph{Advances in neural information processing systems}, pages
  5998--6008.

\bibitem[{Wang et~al.(2019)Wang, Fan, Niu, Yang, Zhang, and Guo}]{Wang2019}
Pengfei Wang, Yu~Fan, Shuzi Niu, Ze~Yang, Yongfeng Zhang, and Jiafeng Guo.
  2019.
\newblock {H}ierarchical {M}atching {N}etwork for {C}rime {C}lassification.
\newblock In \emph{Proceedings of the 42nd International {ACM} {SIGIR}
  Conference on Research and Development in Information Retrieval, {SIGIR}
  2019, Paris, France, July 21-25, 2019}, pages 325--334.

\bibitem[{Wolf et~al.(2020)Wolf, Debut, Sanh, Chaumond, Delangue, Moi, Cistac,
  Rault, Louf, Funtowicz, Davison, Shleifer, von Platen, Ma, Jernite, Plu, Xu,
  Le~Scao, Gugger, Drame, Lhoest, and Rush}]{wolf-etal-2020-transformers}
Thomas Wolf, Lysandre Debut, Victor Sanh, Julien Chaumond, Clement Delangue,
  Anthony Moi, Pierric Cistac, Tim Rault, Remi Louf, Morgan Funtowicz, Joe
  Davison, Sam Shleifer, Patrick von Platen, Clara Ma, Yacine Jernite, Julien
  Plu, Canwen Xu, Teven Le~Scao, Sylvain Gugger, Mariama Drame, Quentin Lhoest,
  and Alexander Rush. 2020.
\newblock Transformers: {S}tate-of-the-{A}rt {N}atural {L}anguage {P}rocessing.
\newblock In \emph{Proceedings of the 2020 Conference on Empirical Methods in
  Natural Language Processing: System Demonstrations}, pages 38--45, Online.
  Association for Computational Linguistics.

\bibitem[{Xiao et~al.(2018)Xiao, Zhong, Guo, Tu, Liu, Sun, Feng, Han, Hu, Wang
  et~al.}]{xiao2018cail2018}
Chaojun Xiao, Haoxi Zhong, Zhipeng Guo, Cunchao Tu, Zhiyuan Liu, Maosong Sun,
  Yansong Feng, Xianpei Han, Zhen Hu, Heng Wang, et~al. 2018.
\newblock {C}ail2018: A large-scale legal dataset for judgment prediction.
\newblock \emph{arXiv preprint arXiv:1807.02478}.

\bibitem[{Xie et~al.(2020)Xie, Ras, van Gerven, and Doran}]{xie2020explainable}
Ning Xie, Gabrielle Ras, Marcel van Gerven, and Derek Doran. 2020.
\newblock {E}xplainable deep learning: {A} field guide for the uninitiated.
\newblock \emph{arXiv preprint arXiv:2004.14545}.

\bibitem[{Xu et~al.(2020)Xu, Wang, Chen, Pan, Wang, and
  Zhao}]{xu-etal-2020-distinguish}
Nuo Xu, Pinghui Wang, Long Chen, Li~Pan, Xiaoyan Wang, and Junzhou Zhao. 2020.
\newblock {D}istinguish {C}onfusing {L}aw {A}rticles for {L}egal {J}udgment
  {P}rediction.
\newblock In \emph{Proceedings of the 58th Annual Meeting of the Association
  for Computational Linguistics}, pages 3086--3095, Online. Association for
  Computational Linguistics.

\bibitem[{Yang et~al.(2019{\natexlab{a}})Yang, Jia, Zhou, and Luo}]{Yang2019}
Wenmian Yang, Weijia Jia, Xiaojie Zhou, and Yutao Luo. 2019{\natexlab{a}}.
\newblock {L}egal {J}udgment {P}rediction via {M}ulti-{P}erspective
  {B}i-{F}eedback {N}etwork.
\newblock In \emph{Proceedings of the Twenty-Eighth International Joint
  Conference on Artificial Intelligence, {IJCAI} 2019, Macao, China, August
  10-16, 2019}, pages 4085--4091.

\bibitem[{Yang et~al.(2019{\natexlab{b}})Yang, Dai, Yang, Carbonell,
  Salakhutdinov, and Le}]{yang2019xlnet}
Zhilin Yang, Zihang Dai, Yiming Yang, Jaime Carbonell, Russ~R Salakhutdinov,
  and Quoc~V Le. 2019{\natexlab{b}}.
\newblock {XLN}et: {G}eneralized autoregressive pretraining for language
  understanding.
\newblock In \emph{Advances in neural information processing systems}, pages
  5753--5763.

\bibitem[{Yang et~al.(2016)Yang, Yang, Dyer, He, Smola, and
  Hovy}]{yang2016hierarchical}
Zichao Yang, Diyi Yang, Chris Dyer, Xiaodong He, Alex Smola, and Eduard Hovy.
  2016.
\newblock Hierarchical attention networks for document classification.
\newblock In \emph{Proceedings of the 2016 conference of the North American
  chapter of the association for computational linguistics: human language
  technologies}, pages 1480--1489.

\bibitem[{Ye et~al.(2018)Ye, Jiang, Luo, and
  Chao}]{ye-interpretableChargePrediction}
Hai Ye, Xin Jiang, Zhunchen Luo, and Wenhan Chao. 2018.
\newblock {I}nterpretable {C}harge {P}redictions for {C}riminal {C}ases:
  {L}earning to {G}enerate {C}ourt {V}iews from {F}act {D}escriptions.
\newblock In \emph{Proceedings of the 2018 Conference of the North {A}merican
  Chapter of the Association for Computational Linguistics: Human Language
  Technologies, Volume 1 (Long Papers)}, pages 1854--1864, New Orleans,
  Louisiana. Association for Computational Linguistics.

\bibitem[{Zeiler and Fergus(2014)}]{zeiler2014visualizing}
Matthew~D Zeiler and Rob Fergus. 2014.
\newblock Visualizing and understanding convolutional networks.
\newblock In \emph{European conference on computer vision}, pages 818--833.
  Springer.

\bibitem[{Zhong et~al.(2018)Zhong, Guo, Tu, Xiao, Liu, and
  Sun}]{LJPTopologicalLearning}
Haoxi Zhong, Zhipeng Guo, Cunchao Tu, Chaojun Xiao, Zhiyuan Liu, and Maosong
  Sun. 2018.
\newblock {L}egal {J}udgment {P}rediction via {T}opological {L}earning.
\newblock In \emph{Proceedings of the 2018 Conference on Empirical Methods in
  Natural Language Processing}, pages 3540--3549, Brussels, Belgium.
  Association for Computational Linguistics.

\bibitem[{Zhong et~al.(2020)Zhong, Wang, Tu, Zhang, Liu, and
  Sun}]{Zhong_Wang_Tu_Zhang_Liu_Sun_2020}
Haoxi Zhong, Yuzhong Wang, Cunchao Tu, Tianyang Zhang, Zhiyuan Liu, and Maosong
  Sun. 2020.
\newblock Iteratively {Q}uestioning and {A}nswering for {I}nterpretable {L}egal
  {J}udgment {P}rediction.
\newblock \emph{Proceedings of the AAAI Conference on Artificial Intelligence},
  34(01):1250--1257.

\end{thebibliography}

\clearpage
\newpage
\appendix

%{\centering\section*{\hfil Supplement\hfil }} \label{supp}
%{\centering\section*{\hfil {\huge\selectfont{Appendix}}\hfil }} \label{supp}

\section*{{\Large\selectfont{Appendix}}} \label{supp}

% \section{Annotations and Case studies}\label{supp:case-study}

\section{Annotations and Case studies: Agreement in Judgment Prediction for Annotators}
\label{supp:case-prediction}
%Due to space constraints, we show only two cases in here.
\noindent {\it \textbf{Annotation Assignment 1954\_13:}} In this case, although the original decision is that the appeal has been rejected, Experts 1-4 have reached the decision that it has been accepted, while Expert 5 has decided that it has been rejected. This discrepancy appears to owe its origin to the very nature of the case and the issues considered by the court. There had been more than one such issue and separate arguments had been made by appellant in favour of each of such issue and associated prayer. The court appears to have agreed to some of the arguments and disagreed with the rest. \\ %Given the binary nature of the choice available to the Users, they could therefore decided the matter either way, as seems to have been done. The exact decision might therefore have been better described as partially accepted/partially rejected, however, this alternative option had not been made available. This indicates that for cases involving multiple issues, providing such a third option might yield more accurate results and uniformity of user-decision. \\
\noindent {\it \textbf{Annotation Assignment 1961\_417:}} In this case, although the original decision is that the appeal has been rejected, Experts 2 and 4 have decided that it has been accepted. Expert 2 appears to have misconstrued certain positions of law and relied unduly upon one of the other cases being cited as precedent (but not considered relevant by the Supreme Court), which might account for the divergence. In case of Expert 4, however, the issue appears to be more of a linguistic matter. Expert 4 has referred to a particular statement made by the court, “The main question that arises in this appeal is whether an illegitimate son of a sudra vis-a-vis his self acquired property, after having succeeded to a half share of his putative fathers estate, will be entitled to succeed to the other half share got by the widow, after the succession opened out to his putative father on the death of the said widow.” From this sentence, Expert 4 has drawn the inference that the appellant was the one asking to establish such entitlement. Since the court in subsequent comments agreed that such entitlement does exist, Expert 4 inferred that the appeal had been accepted. However, in reality, the appellant had been contesting such entitlement. %The court’s singular way of drafting the issue under contention, together with User 4 not having studied more thoroughly the facts leading to the case in the first place (which might have helped the user to reach a conclusion about the appellant’s prayers more accurately) appear to have contributed to this discrepancy.\\        
\noindent {\it \textbf{Annotation Assignment 1962\_47:}} In this case, although the original decision is that the appeal has been rejected, Experts 2 and 5 have decided that it has been accepted. This discrepancy appears to owe its origin to both of them having been misled by Sentence 17 of the case, which appears to refer to the Supreme Court having accepted an appeal and merely giving reasons for such order in the present case. However, the case in point was actually arising from an application for review of the court’s earlier judgment (acceptance of the appeal), and therefore, when the court was affirming its earlier judgment and giving reasons behind it, it was in reality rejecting this present application for review, that had been made by the party (respondent in the original appeal) aggrieved by the acceptance of such appeal by the court earlier. Experts 2 and 5 could not apparently distinguish the appeal from the review petition and that appears to have led to such discrepancy. \\

\section{Models Details} \label{supp:model-details}
Table \ref{tab:results-full} summarizes hyperparameter settings for all the models. All the experiments were run on Google Colab\footnote{\url{https://colab.research.google.com/}} and used the default single GPU Tesla P100-PCIE-16GB, provided by Colab. 
\subsection{Case Prediction Model Details}
\label{supp:cjpe-case-prediction}

%\subsubsection{Classical Models}\label{supp:classical}
\noindent \textbf{Classical Models: } We considered classical ML models like Logistic Regression, SVM, and Random Forest. We used sentence embeddings  via Sent2Vec \cite{pagliardini-etal-2018-unsupervised} and document embeddings via Doc2Vec \cite{le2014distributed} as input features. Both embeddings were trained on \corpuslarge as our data is domain-specific.  Legal proceedings are typically long documents, we tried out extractive summarization methods (as described in \newcite{bhattacharya2019comparative}) for gleaning relevant information from the documents and passing these as input to neural models. However, this approach also resulted in classifiers that were no better than random classifier.

We also experimented by using TF-IDF vectors with the classical models like Logistic Regression (LR), Random Forests (RF) and Support Vector Machines (SVM) from the scikit-learn library in python \cite{scikit-learn}. However, the results were no better than a random classifier, which, according to us, could be due to the huge length of the documents and they were not able to capture such long term dependencies well enough. %Throughout this paper we will call this phenomenon as ``no-learning" zone.
\textbf{Results}: Classical models based on logistic regression and Sent2Vec embeddings performed much worse than the one based on Doc2vec embeddings. It is interesting to see that Doc2Vec+LR has performance competitive to Sequential models. The simple word embedding based model has similar performance as the more complicated hierarchical attention network model (HAN).
% Increasing the dimension size helped in the case of Doc2Vec. However, in Sent2Vec, the best results were seen for dimension size 500. Changing the vocabulary size and min-count during training did not affect the results significantly. Average pooling worked the best for Sent2Vec embeddings. For the hyper-parameters of classical models, we varied the number of estimators for Random Forests and maximum number of iterations for Logistic Regression. However, no significant changes were noticed.
% The results reported in Table \ref{tab:results} are for Doc2Vec and Sent2Vec with LR for max-iterations 50 and 100 respectively. In case of SVM, changing kernel type affected the accuracy by one or half a percent.
The best results are recorded in the Table \ref{tab:results}, each for Sent2Vec and Doc2Vec.

\noindent \textbf{Sequential Models:}
%\label{supp:sequential}
We experimented with standard BiGRU (2 layers) with attention model. We tried 3 different types of embeddings: (i)~Word level trained GloVe embeddings \cite{pennington2014glove}, with last 512 tokens as input, (ii)~Sentence level embeddings (Sent2Vec), where last 150 sentences were input\footnote{last 150 sentences covered around 90\% of the documents}, and (iii)~Chunk level embeddings (trained via Doc2Vec). Both Sequential models and HAN were trained on both \corpuslarge and \corpussmall. All the models from here on were trained on Colab\footnote{\url{https://colab.research.google.com/}}.

% \noindent \textbf{Hyper-parameters:} %\AM{where ever applicable: write about optimizer used and learning rate and any learning schedule followed for all models}\RS{Done}
% For training and evaluation, we used both the \corpuslarge and the \corpussmall datasets.
% \begin{itemize}
%     \item BiGRU:  We experimented with GloVe (last 512 tokens), Sent2Vec (last 150 sentences) and Doc2Vec embeddings.  We tried with both 2 and 3 layers of BiGRU. We chose the hidden dimensions as 100 or 200 and varied epochs from 1 to 5.
%     \item BiGRU + Attention: We experimented with GloVe and Sent2Vec embeddings. We tried with both 2 and 3 layers of BiGRU.  We chose the hidden dimensions as 100 or 200 and varied epochs from 1 to 5.
%     \item HAN: We used the GloVe embeddings to train the model. We used the last 40 sentences with a maximum sentence length of 50. We used a single layered BiGRU for the word as well as the sentence encoder and experimented with hidden dimension of 100 and 200.
% \end{itemize}

We extracted catchphrases \cite{mandal2017automatic}  from the \corpussmall (we could not use this method on \corpuslarge due to requirement of huge compute resources). After extracting these catchphrases we ranked the sentences from the documents accordingly and used upto 200 sentences only\footnote{These covered more than 90\% of the \corpussmall.}. These top 200 sentences were then mapped to their Sent2Vec embeddings and passed through BiGRU as above.\\
% We used Adam as optimizer and a learning rate of $0.001$ for all the three models. For BiGRU and BiGRU + Attention, we used BinaryCrossEntropy as the loss function whereas for HAN we used CategoricalCrossEntropy.

\textbf{Results:} Sequential models trained on \corpuslarge and \corpussmall have similar performances. We also experimented with extracting key sentences from \corpussmall documents with the help of catchphrases and using these sentences as input (via the Sent2Vec embeddings) to a sequence model. Extracting the key sentences performs better than the using all the sentences but the performance is worse (61\% versus 64\% F1) than using GloVe embeddings on last 512 words. GloVe embeddings with BiGRU and attention model gave the best performance (64\% F1) among the sequential models. The GloVe embeddings (last 512 tokens) with BiGRU + Attention gave the best results among the models mentioned above. 
% Increasing the number of layers from 2 to 3 or changing the hidden dimension (like 100 and 200) did not cause any significant variation in results.\\

\noindent\textbf{Transformer Models: } Recently, SOTA language models have been developed using Transformer Architectures \cite{vaswani2017attention}. A number of transformer architectures have been introduced recently. We experimented with BERT \cite{devlin-etal-2019-bert}, DistilBERT \cite{sanh2019distilbert}, RoBERTa \cite{liu2019roberta}, and XLNet \cite{yang2019xlnet}. We used HuggingFace library \cite{wolf-etal-2020-transformers} to fine tune BASE models of above transformers from HuggingFace \cite{wolf-etal-2020-transformers} on the last 512 tokens of \corpuslarge\footnote{As shown in Table \ref{tab:results}, we also experimented with different sections of documents and we observed last 512 tokens gave the best performance}. Due to high compute requirements we could not utilize Longformer \cite{beltagy2020longformer} and Reformer \cite{Kitaev2020Reformer:} models developed especially for long documents.

For the other transformer models we used only the last 512 tokens as input.
% DistilBERT and RoBERTa were trained for 5 epochs, whereas XLNet was trained for 3 epochs (similar as in BERT). For DistilBERT the learning rate was kept the same as the BERT model but for XLNet and RoBERTa we used a learning rate of $2e-6$. We kept the batch size as 6 in these transformers as well.

\textbf{Results: }Among the combinations of input tokens, the best performance was obtained by using last 512 tokens as input to the BERT Base model. We can observe the trend that the more the tokens from the final parts of the document are taken as input, the better is the prediction performance. This observation agrees with the fact that there are more clues towards the correct prediction in the final parts of the document (since {\it Arguments}, {\it Ratio of the decision} etc. \newcite{bhattacharya2019identification} most aligned to the judgment  are expected to appear more towards the end, closer to the judgment). As for the comparison between different transformers, unsurprisingly, RoBERTa and XLNet perform better than BERT in the prediction sub-task. Similarly, among DistilBERT and BERT, the latter outperforms the other.

% \begin{figure*}[!h]
%     \centering
%     \frame{\includegraphics[width=0.9\textwidth]{images/occal25_new.pdf}}
%     \caption{Visualization of Occlusion scores accross full Test set.}
%     \label{fig:occ_scores_all}
% \end{figure*}

% \begin{figure*}[!h]
%     \centering
%     \frame{\includegraphics[width=0.9\textwidth]{images/attall25_new.pdf}}
%     \caption{Visualization of Attention scores accross full Test set.}
%     \label{fig:att_scores_all}
% \end{figure*}

\begin{figure*}[b]
    \centering
    \frame{\includegraphics[width=\linewidth]{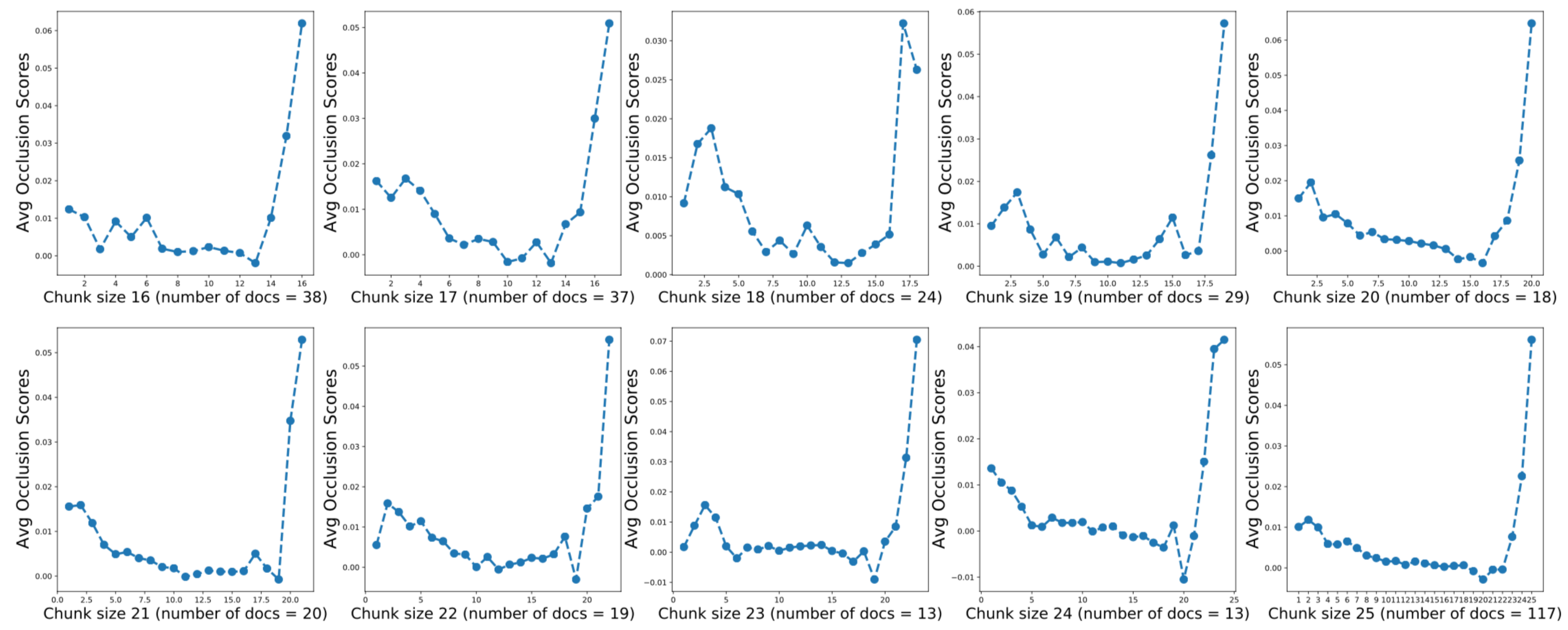}}
    \caption{Visualization of Occlusion scores accross full Test set.}
    \label{fig:occ_scores_all}
\end{figure*}

\begin{figure*}[b]
    \centering
    \frame{\includegraphics[width=\linewidth]{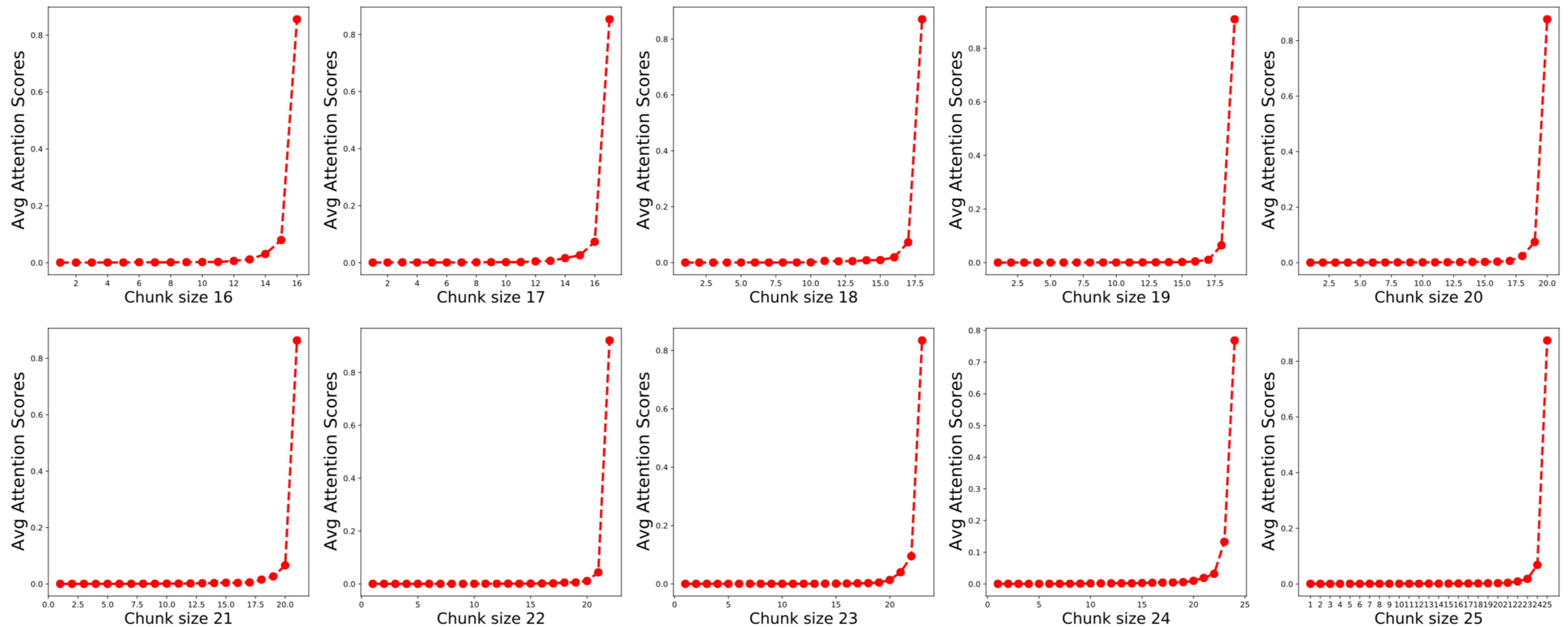}}
    \caption{Visualization of Attention scores accross full Test set.}
    \label{fig:att_scores_all}
\end{figure*}

\noindent\textbf{Hierarchical Models: } %\label{sup:hmodel}
In order to use transformers hierarchically, it was first necessary to fine-tune these models on the downstream task of classification. We use two different strategies to fine-tune these:
\begin{itemize}
    \item On \corpuslarge: Using last 512 tokens only from the documents.
    % \item On \corpuslarge: We fine-tune the transformer using last 512 tokens only from the documents, similar to as what we did in the transformers section.
    \item On \corpussmall: We fine-tune the transformer by dividing each document into chunks of 512 with an overlap of 100 tokens, the label for each chunk is given as the whole document label.
\end{itemize}
Then we extracted the 768 dimension, $[CLS]$ token embeddings from the transformers for each chunk in all the documents. This was done on \corpuslarge corpus irrespective of whether it was fine-tuned on \corpuslarge or \corpussmall. As mentioned in \cite{devlin-etal-2019-bert} we also experimented with concatenating the last 4 hidden layers of the $[CLS]$ token and taking that as the chunk embedding. \\
% \begin{figure}
% \hskip-1cm
%     \centering
%     \includegraphics[scale=0.25]{images/architecture.pdf}
%     %\includegraphics[width=\linewidth,height=7cm]{images/architecture.pdf}
%     \vspace{-1mm}
%     \caption{Hierarchical XLNet architecture (XLNet + BiGRU)} %\AB{Figure size is too small}
%     \label{fig:xlnet}
%     \vspace{-7mm}
% \end{figure}
After getting the chunk embeddings we used two types of neural networks: BiGRU and CNN.
% \begin{itemize}
%     \item BiGRU: We experimented with 2 and 3 layers of BiGRU by varying epochs from 1 to 5. Due to limitations of BiGRU, we limited the input to the last 10,000 tokens of the document\footnote{Last 10,000 tokens captured 90\% of documents}. We also experimented with adding attention layer on the top similar as in BiGRU + attention in sequential models.
%     \item CNN: We experimented with this method only for the models fine-tuned on \corpuslarge as they were giving the best results.
% \end{itemize}
% Overall the training of some of these models was quite unstable. We used only one epoch for some models as they were over-fitting if the training was continued. 

For some models, the results varied over multiple runs. For these we recorded their mean and variance on F1 and Accuracy in the \cref{tab:results}.

\textbf{Results: }Information is lost in considering only the last portion of the case proceeding for prediction and this is reflected in the performance of hierarchical models. In general, all the hierarchical models outperform transformer models. Adding attention on top of BiGRU in the hierarchical model does not boost the performance significantly. However, adding a CNN (instead  of BiGRU + Attention) on top gives a competitive performance. As for the comparison between the strategies of fine-tuning between \corpuslarge and \corpussmall, the later seemed to perform worse on prediction. For the hierarchical concatenated model fine tuned on \corpussmall, there was a slight boost in performance. 

% \begin{figure}
%     \centering
%     \includegraphics[width=11cm, height=8cm]{images/OverlapMax.png}
%     \caption{Caption}
% \end{figure}

%%%%%%%%%%%%%%%%%%%%%%%%%%%%%%%%%%%%%%%%%%%%%%%%%%%%%%%%%
%%%%%%%%%%%%%%%%%%%%%%%%%%%%%%%%%%%%%%%%%%%%%%%%%%%%%%%%%
%%%%%%%%%%%%%%%%%%%%%%%%%%%%%%%%%%%%%%%%%%%%%%%%%%%%%%%%%
%%%%%%%%%%%%%%%%%%%%%%%%%%%%%%%%%%%%%%%%%%%%%%%%%%%%%%%%%
%%%%%%%%%%%%%%%%%%%%%%%%%%%%%%%%%%%%%%%%%%%%%%%%%%%%%%%%%
%%%%%%%%%%%%%%%%%%%%%%%%%%%%%%%%%%%%%%%%%%%%%%%%%%%%%%%%%

\subsection{Explanability Models and Results Details}
\label{supp:explain}

% \begin{figure}
%      \begin{subfigure}[b]{0.5\textwidth}
%          \centering
%          \includegraphics[width=\linewidth]{images/occpdf.pdf}
%          \caption{Occlusion}
%          \label{fig:y equals x}
%      \end{subfigure}
%      \begin{subfigure}[b]{0.5\textwidth}
%          \centering
%          \includegraphics[width=\linewidth]{images/attpdf.pdf}
%          \caption{Attention}
%          \label{fig:three sin x}
%      \end{subfigure}
%         \caption{Visualizing chunk scores}
%         \label{fig:attention_occlusion_graphs}
% \end{figure}
% \begin{figure*}[t]
%     \centering
%     \frame{\includegraphics[width=\textwidth]{images/bertviz.pdf}}
%     \caption{Token visualization heatmap obtained using Integrated Gradients (via Captum library) for document name 1951\_33.txt in \corpusexpert, for the BERT model. Green color corresponds to the tokens that contributed positively towards the prediction, whereas Red color corresponds to the tokens that have negative contributions.}
%     \label{fig:my_label}
% \end{figure*}

To extract explanations from our best model (XLNet + BiGRU), we propose a method inspired from  \newcite{li2016understanding} and \newcite{zeiler2014visualizing}. The idea is to use occlusion method at both levels of the hierarchy. For the BiGRU part of the model, for each document we mask each complete chunk embedding one at a time. The masked input is passed through the trained BiGRU and output probability (masked probability) of the label obtained by original unmasked model is calculated. The masked probability is compared with unmasked probability to calculate chunk explainability score. Formally, for a chunk $c$, if the sigmoid outputs (of the BiGRU) are $\sigma_{m}$ (when the chunk was not masked) and $\sigma_{m'}$ (when the chunk was masked) and the predicted label is ${y}$  then the probabilities and chunk score 
$s_{c} = p_{m} - p_{m'}$ and $p_{m'/m} =\begin{cases}
			\sigma_{m'/m}, & \hspace{-2mm} y = 1\\
            1 - \sigma_{m'/m}, & \hspace{-2mm} y = 0
		 \end{cases}$

We obtain sentences that explain the decision from the transformer part of the model (XLNet) using the chunks that were assigned positive scores. Each chunk (length 512 tokens) is segmented into sentences using NLTK sentence splitter \cite{loper2002nltk}. Similar to BiGRU, each sentence is masked and the output of the transformer at the classification head (softmax logits) is compared with logits of the label corresponding to original hierarchical model. The difference between the logits normalized by the length of the sentence is the explanation score of the sentence. Finally, top-k sentences ($\sim$ 40\%) in each chunk are selected.

%\AM{Write about token visualizations}
% Our models are implemented in PyTorch while the original DeepLIFT code is implemented in CAPTUM library \cite{captum2019github}. However, CAPTUM does not support DeepLIFT methods on RNNs in PyTorch yet. Moreover, CAPTUM does not have any implementation for LRP. It is mentioned that they cannot access the building blocks of PyTorch’s built-in LSTM, RNNs and GRUs such as Tanh and Sigmoid.
% As mentioned in \cref{sec:explanation-model}, we visualized the attributions calculated by using Integrated Gradients over the BERT base transformer that was trained over last 512 tokens only. Due to computational limits, we restricted the parameter of $n\_steps$ in the integrated gradients to 50 only. In Figure \ref{fig:my_label}, we provide a visualization of some sentences from one of the $56$ documents taken from \corpusexpert. It shows both positively and negatively contributing tokens towards the prediction. The true label for the case is ``Rejected''. The model correctly highlights the importance to the words like `evidence', `defendants' and `plaintiff' in the sentences.

% However, if we see the explanations from the expert annotators (Table \ref{tab:annotator_explanation}), it is clear that we cannot conclude the decision of a court case using token level importance only. These tokens may be contributing towards the correct prediction for the machine learning model but these cannot function as the explanations of the prediction for humans.

In Figure \ref{fig:occ_scores_all} and Figure  \ref{fig:att_scores_all} we visualize the mean chunk importance scores. Out of the 1517 test documents we average out chunk scores of the documents having same number of chunks. As shown in Figure \ref{fig:att_scores_all}, the attention weights are biased towards the last chunks, thus giving negligible attention to the chunks before. However, in Figure \ref{fig:occ_scores_all}, in some of the graphs, the last chunk is given the second-highest score and in 7 out of 10 graphs, it has the highest score. 
% In rows 2 and 3 of the Figure  \ref{fig:occ_scores_all}, we can see how the pattern matches with that of the attention weights given. 
Due to space limitation, we are not providing the graphs for occlusion and attention scores for chunks 1 to 15. But we observed that for these chunks pattern matches for occlusion scores with attention scores. From these observations, we believe it is safe to say that both the methods of visualization affirm our hypothesis that \textit{the most relevant syntactic and semantic information lies towards the end of the case.}
Although attention scores are optimized (via loss minimization or accuracy maximization) to concentrate on last chunks, this is not the case with occlusion scores. There is no optimization of occlusion scores, yet they still focus on the chunks at the end which affirms our hypothesis. One might argue that this observation might be due to the transformer being trained on last 512 tokens only. To check this, we also visualized the hierarchical transformers trained on \corpussmall, but the results were similar as to what we have observed in this case.

\begin{table}[!h]
%\begin{center}
% \small
%\hski-2.1cm
\vspace{-3.56cm}
\resizebox{\columnwidth}{!}{
\begin{tabular}{|l|l|}
\hline
%%%%%%%%%%%%%%%%%%%%%%%%%%%%%%%%%%%%%%%%%%%%%%%%%%%%
\textbf{Model}& \textbf{\begin{tabular}[c]{@{}l@{}}Hyper-Parameters (E = Epochs), \\ (Dim = Embedding Dimension), \\ (L = Layers), (att. = attention),\\ (default setting= 512 tokens with\\ overlapping 100 tokens)\end{tabular}} \\ \hline
%%%%%%%%%%%%%%%%%%%%%%%%%%%%%%%%%%%%%%%%%%%%%%%%%%%%
\multicolumn{2}{|l|}{\textbf{Classical Models on \corpuslarge train set}} \\ \hline
\textbf{Doc2Vec + LR} & dim = 1000 , E = 20 \\\hline
Sent2vec + LR & dim=500, E = 20, Avg Pool \\ \hline
%%%%%%%%%%%%%%%%%%%%%%%%%%%%%%%%%%%%%%%%%%%%%%%%%%%%
\multicolumn{2}{|l|}{\textbf{Sequential Models on \corpuslarge train set}} \\ \hline
Sent2vec + BiGRU + att. & dim = 200, E = 1, L = 2 \\ \hline
Doc2vec + BiGRU + att.  & dim = 1000,  E = 2, L = 2 \\ \hline 
\textbf{GloVe + BiGRU + att.} & dim = 180, E = 3, L = 2 \\ \hline
HAN & \begin{tabular}[c]{@{}c@{}}word dim = 100, sent dim = 100, \\ E = 10 \end{tabular}\\ \hline
%%%%%%%%%%%%%%%%%%%%%%%%%%%%%%%%%%%%%%%%%%%%%%%%%%%%
\multicolumn{2}{|l|}{\textbf{Sequential Models on \corpussmall train set}} \\ \hline
Sent2Vec + BiGRU+ att.  & dim = 200, E = 1, L = 2  \\ \hline
Doc2vec + BiGRU + att.  & dim = 1000,  E = 2, L = 2 \\ \hline
\textbf{GloVe + BiGRU + att.} & dim = 180, E = 10, L = 2 \\ \hline
HAN & \begin{tabular}[c]{@{}c@{}}word dim = 100, sent dim = 100, \\ E = 10 \end{tabular} \\ \hline
\begin{tabular}[c]{@{}c@{}}Catchphrases + Sent2Vec\\  + BiGRU + att.\end{tabular}  & dim =180, E =5, L = 2  \\ \hline
%HAN & X & X  & X & X & X \\ \hline
%%%%%%%%%%%%%%%%%%%%%%%%%%%%%%%%%%%%%%%%%%%%%%%%%%%%
\multicolumn{2}{|l|}{\textbf{Transformer Models on \corpuslarge train set}} \\ \hline
BERT Base & 512 begin tokens, E = 3  \\ \hline
BERT Base & 256 begin, 256 end tokens,  E = 3  \\ \hline
BERT Base  & 256 mid, 256 end tokens, E = 3  \\ \hline
BERT Base  & 128 begin, 256 mid, 128 end,  E = 3 \\ \hline
BERT Base & 512 end tokens,  E = 3  \\ \hline
DistillBERT & 512 end tokens, E = 5  \\ \hline
\textbf{RoBERTa}  & 512 end tokens, E = 5  \\ \hline
XLNet  & 512 end tokens, E = 3  \\ \hline
%%%%%%%%%%%%%%%%%%%%%%%%%%%%%%%%%%%%%%%%%%%%%%%%%%%%
\multicolumn{2}{|l|}{\textbf{Hierarchical Models on \corpuslarge train set}}  \\ \hline
BERT + BiGRU  & default setting, E = 5, L = 3  \\ \hline
RoBERTa + BiGRU & default setting, E = 2, L = 3, runs = 3 \\ \hline
\textbf{XLNet + BiGRU}  & default setting, E = 5, L = 2  \\ \hline
BERT + CNN & default setting, E = 3, L = 3 (Conv1D)  \\ \hline
RoBERTa + CNN  & default setting, E = 3, L = 3 (Conv1D) \\ \hline
XLNet + CNN & default setting, E = 3, L = 3 (Conv1D) \\ \hline
%%%%%%%%%%%%%%%%%%%%%%%%%%%%%%%%%%%%%%%%%%%%%%%%%%%%
\multicolumn{2}{|l|}{\textbf{Hierarchical Models on \corpussmall train set}} \\ \hline
BERT + BiGRU & default setting, E = 1, L = 2, 3 runs \\ \hline
RoBERTa + BiGRU & default setting, E = 1, L = 2, 3 runs \\ \hline
XLNet + BiGRU & default setting, E = 2, L = 2, 3 runs  \\ \hline
%%%%%%%%%%%%%%%%%%%%%%%%%%%%%%%%%%%%%%%%%%%%%%%%%%%%
\multicolumn{2}{|l|}{\textbf{Hierarchical Models with Attention on \corpuslarge train set}} \\ \hline
BERT + BiGRU + att. & default setting, E = 2, L = 2, 3 runs \\ \hline
RoBERTa + BiGRU + att.  & default setting, E = 2, L = 3, 3 runs  \\ \hline
\textbf{XLNet + BiGRU + att.} & default setting, E = 3, L = 2, 3 runs \\ \hline
%%%%%%%%%%%%%%%%%%%%%%%%%%%%%%%%%%%%%%%%%%%%%%%%%%%%
\multicolumn{2}{|l|}{\textbf{Hierarchical Models with Attention on \corpussmall train set}}  \\ \hline
BERT + BiGRU + att. & default setting, E = 1,  L = 2, 3 runs \\ \hline
RoBERTa + BiGRU + att.  & default setting, E = 1,  L = 3, 3 runs \\ \hline
\textbf{XLNet + BiGRU + att.}    & default setting, E = 1,  L = 2, 3 runs \\ \hline
%%%%%%%%%%%%%%%%%%%%%%%%%%%%%%%%%%%%%%%%%%%%%%%%%%%%
\multicolumn{2}{|l|}{\textbf{Transformers Voting Ensemble}} \\ \hline
\textbf{RoBERTa} & fine tuned on last 512 tokens, voting \\ \hline
XLNet & fine tuned on last 512 tokens, voting \\ \hline
%%%%%%%%%%%%%%%%%%%%%%%%%%%%%%%%%%%%%%%%%%%%%%%%%%%%
\multicolumn{2}{|l|}{\textbf{Hierarchical concatenated model with att on \corpussmall train}} \\ \hline
\textbf{XLNet + BiGRU} & \begin{tabular}[c]{@{}c@{}}last 4 layers concat, E = 1,\\ L = 2, 3 runs \end{tabular}\\ \hline
%%%%%%%%%%%%%%%%%%%%%%%%%%%%%%%%%%%%%%%%%%%%%%%%%%%%
\end{tabular}}
%\end{center}
\caption{Hyper-parameters corresponding to every model.} %\SN{We can remove hyper-parameter column and can make separate table for same at appendix. That will save one entire column.} \AB{agree --  we can move}}
\label{tab:results-full}
\end{table}

\end{document}